\documentclass{article} 
\usepackage{collas2025_conference,times}
\usepackage{easyReview}

\usepackage{amsmath,amsfonts,bm}

\def\eqref#1{equation~\ref{#1}}

\def\1{\bm{1}}

\DeclareMathAlphabet{\mathsfit}{\encodingdefault}{\sfdefault}{m}{sl}
\SetMathAlphabet{\mathsfit}{bold}{\encodingdefault}{\sfdefault}{bx}{n}

\usepackage{hyperref}
\hypersetup{
    colorlinks=true,
    linkcolor=red,
    filecolor=magenta,
    urlcolor=blue,
    citecolor=purple,
    pdftitle={Overleaf Example},
    pdfpagemode=FullScreen,
    }

\usepackage{algorithm}
\usepackage{algpseudocode}
\usepackage{subcaption}
\usepackage{multirow}
\usepackage{booktabs}
\usepackage{wrapfig}
\usepackage{mathtools}

\title{Combining Pre-Trained Models for Enhanced Feature Representation in Reinforcement Learning}

\author{Elia Piccoli\textsuperscript{1,2,3}\thanks{Correspondence to: elia.piccoli@phd.unipi.it}, 
        Malio Li\textsuperscript{1}, 
        Giacomo Carfì\textsuperscript{1}, 
        Vincenzo Lomonaco\textsuperscript{1}, 
        Davide Bacciu\textsuperscript{1}\\
        \textsuperscript{1} Department of Computer Science, University of Pisa, Italy \\
        \textsuperscript{2} Department of Computing Science, University of Alberta, Canada $\;$ \textsuperscript{3} Amii
}

\collasfinalcopy 

\begin{document}

\maketitle

\begin{abstract}
The recent focus and release of pre-trained models have been a key components to several advancements in many fields (e.g. Natural Language Processing and Computer Vision), as a matter of fact, pre-trained models learn disparate latent embeddings sharing insightful representations. On the other hand, Reinforcement Learning (RL) focuses on maximizing the cumulative reward obtained via agent's interaction with the environment. RL agents do not have any prior knowledge about the world, and they either learn from scratch an end-to-end mapping between the observation and action spaces or, in more recent works, are paired with monolithic and computationally expensive Foundational Models. How to effectively combine and leverage the hidden information of different pre-trained models simultaneously in RL is still an open and understudied question. In this work, we propose Weight Sharing Attention (WSA), a new architecture to combine embeddings of multiple pre-trained models to shape an enriched state representation, balancing the tradeoff between efficiency and performance. We run an extensive comparison between several combination modes showing that WSA obtains comparable performance on multiple Atari games compared to end-to-end models. Furthermore, we study the generalization capabilities of this approach and analyze how scaling the number of models influences agents' performance during and after training.
\end{abstract}

\section{Introduction}\label{sec:introduction}
Starting from the first pioneering works of \citet{mnih2013playing, mnih2015human}, which started to explore the use of deep neural networks in games, such as Arcade Learning Environment (ALE) \citep{bellemare2013atari}, a completely different perspective and scenario has been unleashed. As a matter of fact, an increasing interest has been gathering towards Deep Reinforcement Learning (DRL). DRL has produced substantial advancements in various research fields and applications, such as games, beating all Atari games at super-human performance with a single algorithm \citep{agent57} or defeating pro players in games like GO, StarCraft II and DOTA \citep{alphago, starcraft, dota}, but also in robotics \citep{rlrob, bousmalis2023robocat} and lastly toward general agents for 3D environments including different data modalities \citep{sima2024}.

In a typical RL setting, agents receive as input the raw representation of the state, without any additional information on the elements that characterize it. The learning process, focused on creating a policy - a mapping from states to actions that determines agent behavior - implicitly hides a significant effort related to understanding how to process the input representations. While one of the promises of deep learning algorithms is to automatically construct well-tuned features, such representation might not emerge from end-to-end training of deep RL agents. Agents learn how to solve the task while indirectly learning how to process and extract useful information from the input features. Although this approach has been the solution for several works, it adds an additional layer of complexity over RL algorithms, eventually requiring significant computational resources.

Several works analyze the differences between the learning process of humans and RL agents, highlighting how prior knowledge can influence the learning curve \citep{lake2017building, dubey2018investigating}. Humans and agents define the learning pattern by incrementally improving their abilities. Human learning often involves incremental skill development, leveraging prior knowledge, and adapting strategies based on experience.
In new environments with few interactions, humans can leverage their prior abilities to understand the effects of their actions. This allows them to focus mostly on learning a good policy, rather than also learning how to interpret the environment. How can we encode and represent abilities in RL agents?

In light of this question, pre-trained models present themselves as good candidates. They are trained on extensive datasets in a self-supervised fashion, enabling them to learn rich and generalizable representations. These representations can be leveraged in RL to accelerate the learning process and improve performance on complex tasks.
The objective of this paper is to study how different latent embeddings can be \texttt{combined} to compute a rich and informative representation of the environment.
In this way, we can focus on \texttt{how} RL agents can exploit prior knowledge provided by multiple pre-trained models and improve the learning process of the actual policy. Here we report the main contributions of this work:

\begin{itemize}
    \item We propose Weight Sharing Attention (WSA) as a combination module to incorporate different encodings coming from several pre-trained models. This approach achieves comparable performance with end-to-end solutions while also adding robustness and dynamic scalability to RL agents.
    \item We run experiments across multiple Atari games studying the behavior of 7 different combination modules. To the best of our knowledge, this is the first analysis focused on the combination of multiple pre-trained models' latent representation in RL.
    \item We show that without any fine-tuning of hyperparameters, or exploiting a small search, our methodology can achieve comparable results with established RL end-to-end solutions. Moreover, we prove and address the problems of \textit{distribution shifts}, \textit{failing robustness}, and \textit{adaptation to evolving knowledge}, which presents themselves as the key challenges for this compositional RL approach.
\end{itemize}

\section{Related Work}\label{sec:related_work}
The idea of improving RL agents feature representation by providing information already processed by other models is not new. Some work focus on feature representations that can be detached from the learning process by having a module that is able to extract relevant information. Agents learn how to perform tasks by integrating different enriched elements, rather then needing to learn the state space from scratch. Some works use human-defined data \citep{bramlage2022generalized} or leverage pre-trained models' feature representation learned on different tasks, for example, computer vision \citep{shah2021rrl, blakeman2022selective, yuan2022pre} or other RL tasks \citep{kumar2022offline}. Several approaches learn a general representation from collected data by applying different self-supervised methodologies, which can also be fine-tuned during policy learning \citep{goel2018unsupervised, anand2019unsupervised, kulkarni2019unsupervised, schwarzer2021pretraining, xiao2022masked, montalvo2023exploiting, majumdar2023avc}. In order to be effective, these representations should be as general as possible, without having any bias with respect to the task the agent should solve.
Other works try to overcome this problem by presenting methodologies to build representations that are reward free - meaning they are unbiased with respect to the task \citep{stooke2021decoupling} - or by adding auxiliaries objectives to help shape the latent representation \citep{lan2023bootstrapped}. Despite achieving good results in model-free RL, \citet{failpre-train-mb} studies the limitations of using one single model to gather pre-trained visual features in model-based RL. Several state-of-the-art methodologies and works are employing Mixture of Experts to their models, suggesting that combining multiple sources of information improve the performance \citep{moerl}. However, \textit{how} to combine multiple latent features in a single representation is still an open problem. There is almost non-existing literature about works that try to apply more than one model at the same time to enhance the representation. \citet{sima2024} provide the agent with multiple instances of pre-trained models that solve auxiliary tasks and combine their information. They showcase that combining pre-trained models with trained-from-scratch components leads to better-performing agents. Unfortunately, due to the large scale of their work, there is no particular attention towards how models' representations are combined. Recent advances in large-scale language models have increased the research community's interest in combining visual information with textual data. In these scenarios, agents must handle different data modalities at the same time. This can be obtained by using either a single multi-modal pre-trained model \citep{liu2022instruction} or multiple models for each data source. A recent work that follows the latter idea is OpenVLA \citep{kim2024openvla}. In this model visual representations are embedded using multiple vision encoders. The embeddings are then combined and projected, using a small MLP, into a common latent space with the language model. In our work, we aim to focus specifically on how to combine pre-trained models. In this context, our work could be used as a component within the aforementioned works to improve the effectiveness of combining different pre-trained models that possibly operate across different data types. We propose WSA as an effective and efficient solution to merge different pre-train models' embeddings while guaranteeing the ability to scale over time.

\begin{figure}[ht]
    \begin{center}
        \includegraphics[width=0.9\textwidth]{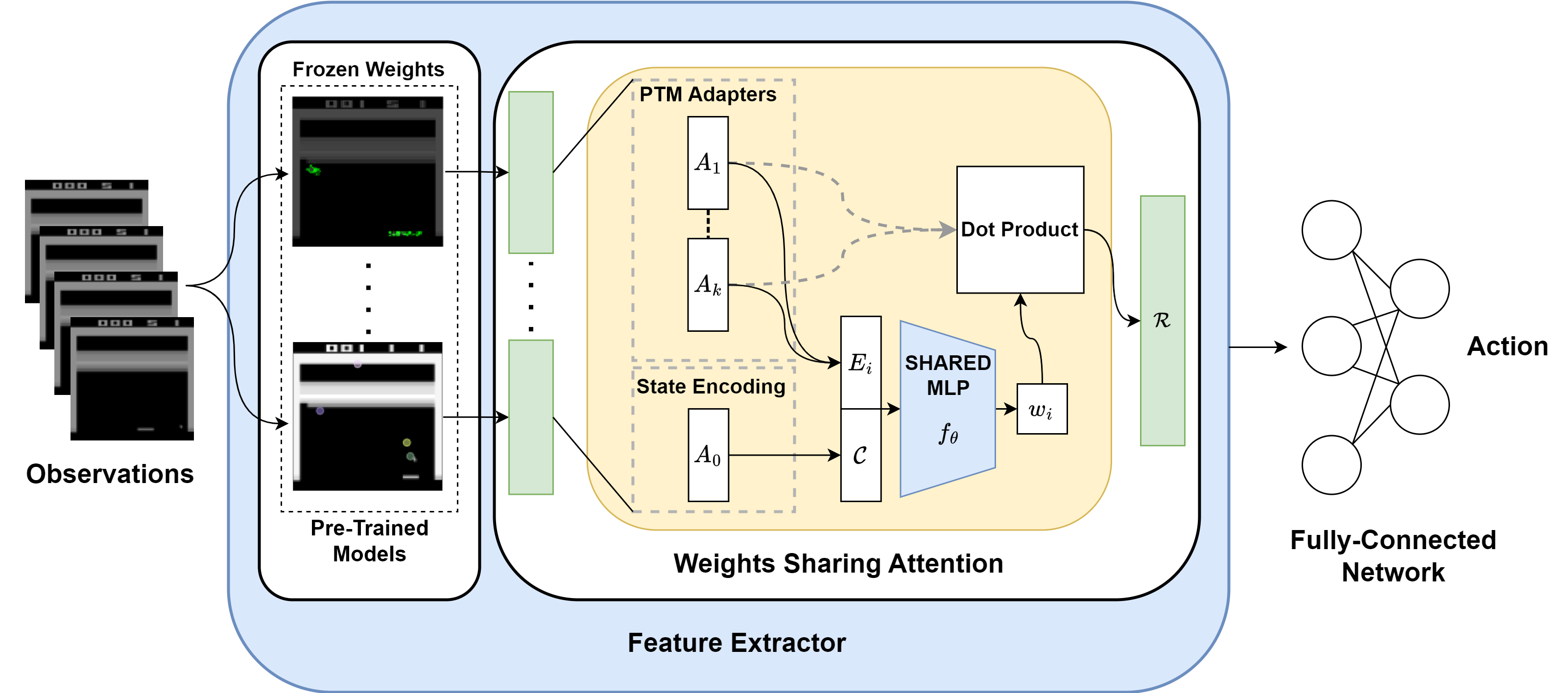}
    \end{center}
    \caption{A schematic representation of the main pipeline from observations to actions and WSA architecture. The last four frames are stacked together and passed as input to the pre-trained models and their adapters $\mathcal{A}_1, \dots \mathcal{A}_k$ to obtain the embeddings of the current state. A State Encoder $\mathcal{E}$ and its adapter $\mathcal{A}_0$ compute an encoding of the state, that will be used as context $\mathcal{C}$. A shared MLP layer takes as input the context $\mathcal{C}$ and the representation of the $i^{th}$ model, $E_i$, and predicts its weight. The final enriched representation $\mathcal{R}$ is obtained from the weighted summation between each weight $w_i$ and the respective embedding $E_i$. Finally, $\mathcal{R}$ is given as input to agents' \texttt{Fully-Connected Network}.}
    \label{fig:main_architecture}
\end{figure}

\section{Weight Sharing Attention (WSA)}\label{sec:method}
Many previous works share the idea of leveraging pre-trained models to enhance the state representation, allowing world knowledge transfer and simplifying the RL training process. Differently from these studies, our work focuses on how to use multiple models at the same time and how to effectively combine their latent representations to improve agents performance. Our methodology enhances the efficiency and effectiveness of RL agents by leveraging a set of pre-trained models, tailored to the current task, that serve as prior knowledge.

The architecture of our agents is divided into two main components, as shown in Figure \ref{fig:main_architecture}. The first one, delimited by the \textcolor{blue}{blue} box, handles the processing of the current observation into a latent representation. While agents interact with the environment, the current observation is collected and processed through each pre-trained model. This process yields multiple perspectives or views of the world, each encoded uniquely based on the specific model. These features represent different facets of the environment, providing the agent with a comprehensive and enhanced understanding of the current state. The challenge lies in integrating these disparate views into a cohesive and enriched representation. To achieve this, we employ a specialized combination module, \textcolor{orange}{yellow} box, that synthesizes the information from various models into a single latent representation. The intuition is that the enriched representation provides a comprehensive understanding of the current state of the environment, derived from the collective insights of the pre-trained models. By leveraging this information, our approach significantly reduces the burden on the RL agent to learn the environmental representation from scratch, allowing it to focus more on refining the action-mapping process. This accelerates the learning process and enhances overall performance by providing a well-structured and informative representation. In the second part, the unified representation is fed into a small neural network that maps the latent encoding to actions.

To facilitate the integration of different pre-trained models, we propose Weight Sharing Attention (WSA). The WSA module leverages weight sharing and attention principles to merge outputs from diverse models efficiently. The approach can be adapted and extended to any number and type of pre-trained models while maintaining its structure. Figure \ref{fig:main_architecture} outlines the main pipeline and components required by our module, while Algorithm \ref{alg:wsa} reports the pseudocode for the actual implementation.

\begin{algorithm}[ht]
    \caption{Weight Sharing Attention}\label{alg:wsa}
    \texttt{In \textcolor{blue}{blue} we highlight the components updated during training.}
    \begin{algorithmic}[1]
        \Require Observation $\mathcal{O}$, Set of Pre-Trained Models $\Psi$, State Encoder $\mathcal{E}$
        \State $\mathcal{C} = \textcolor{blue}{\mathcal{A}_0}(\mathcal{E}(\mathcal{O}))$ 
        \Comment{Compute context representation $\mathcal{C}$}
        \For {pre-trained model $\psi$ in $\Psi$}
            \State $x = \psi_i(\mathcal{O})$
            \Comment{Forward pass \textit{i-th} pre-trained model using current observation}
            \State $E_i = \textcolor{blue}{\mathcal{A}_i}(x)$
            \Comment{Use the adapter to compute the resized embedding $E_i$}
            \State $w_i = \textcolor{blue}{f_\theta}(\mathcal{C}, E_i)$
            \Comment{Compute the weight $w_i$ using the \texttt{Shared Weight Network} $f_\theta$}
        \EndFor
        \State $W = \frac{W}{\lVert W \rVert_1}$
        \Comment{Normalize weights $W$ using L1-norm}
        \State $\mathcal{R} = \sum_{i=0}^{|\Psi|} w_i * E_i$
        \Comment{Compute final representation: weights $W$, embeddings $E$ $\rightarrow$ $W \cdot E$}
    \end{algorithmic}
\end{algorithm}

Each pre-trained model is equipped with an adapter $\mathcal{A}$ - a linear layer followed by a non-linearity - that maps its latent representation into an embedding $E$ of predefined size. This shared space ensures that different model outputs are compatible and can be seamlessly integrated. Adapters' parameters are optimized during training. Among the provided models, one is used as State Encoder $\mathcal{E}$ to compute a representation of the current state, which serves as context $\mathcal{C}$ for the attention mechanism. A Multi-Layer Perceptron (MLP) $f_\theta$ (\texttt{Shared Weight Network}) receives as input the context and the single embedding of each pre-trained model, and outputs the model-specific weight $w_i$ to determine its contribution in the final representation. This MLP is shared across all models, ensuring uniform weight computation. The weight vector $W$ is normalized by its L1 norm to produce a probability distribution $p = \frac{W}{\lVert W \rVert_1}$, ensuring the resulting values are non-negative and sum to one. By dynamically adjusting weights based on the current context, WSA emphasizes the most relevant models for any given situation, resulting in a more accurate and adaptable representation. The final representation $\mathcal{R}$ is computed using the weights for all pre-trained models. This involves a dot product between the embeddings of the pre-trained models and their corresponding weight, producing a weighted sum that combines various perspectives into a single, enriched latent representation. This final representation is employed to learn a mapping to actions, thereby enhancing the performance and learning efficiency of RL agents by integrating the strengths of multiple pre-trained models.The \texttt{Fully-Connected Network} maps the final embedding $\mathcal{R}$ to actions.

Our approach by design guarantees two additional features. WSA is \texttt{scalable} and \texttt{adaptable} to any number of pre-trained models. Its shared component can handle an arbitrary number of models, making it a flexible solution for dynamically evolving agents. On this matter, Section \ref{sec:scaling} analyzes two different scenarios where the number of pre-trained models is increased or decreased over time. Additionally, WSA adds \texttt{explainability} to the model. By examining the weights assigned to different pre-trained models, we can gain insights into which models are most influential in a given context. This characteristic is important for achieving a more interpretable decision-making process for RL agents and for gaining valuable insights into the learning process.

\section{Experiments \& Results}\label{sec:experiments}
In our experimental evaluation, we investigate the benefits and challenges that raise when combining multiple pre-trained models in RL. Appendix \ref{sec:app-exp-setup} provides a detailed overview of the experimental setup used in this study. In particular, we aim to answer the following research questions:
\begin{itemize}
    \item[(\texttt{Q1})] Does the combination of pre-trained models improve the performance compared to end-to-end architectures?
    \item[(\texttt{Q2})] Are pre-trained models resilient to changes in the environment?
    \item[(\texttt{Q3})] Does learning to combine multiple representations improve agents' robustness to changes in the environment?
    \item[(\texttt{Q4})] What happens when the number of pretrained models changes over time?
\end{itemize}

\paragraph{Pre-Trained Models: data and training.}\label{sec:pre-trainm}
To pretrain the different models, we first create a specific dataset for each game. For each environment, we collect \texttt{1M} frames via random agents interacting with the environment.
During the training process, data is randomly sampled to avoid any correlation between elements due to the collection phase. For the experiments concerning our approach, we use three different models, which provide us with four different pre-trained networks: \texttt{Video Object Segmentation} \citep{goel2018unsupervised}, \texttt{State Representation} \citep{anand2019unsupervised}, and \texttt{Object Keypoints} \citep{kulkarni2019unsupervised}. We implement and train the models for all the games using the default architecture and hyperparameters, the values and additional details, e.g. the embedding representation shape, are reported in Appendix \ref{sec:app-models}. Additionally, we train a deep autoencoder - inspired by Nature CNN \citep{mnih2015human} - to encode the current state and leverage its representation to act as the context $\mathcal{C}$ in our approach. As previously mentioned, the main objective of this work is \texttt{combining} different representations, thus, we exploit these models, which can be easily trained and adapted compared to monolithic foundational models but still provide insightful and useful information. Nevertheless, the chosen pre-trained models can be arbitrarily changed to more complex or larger ones without affecting the structure of our work. While training the agent, pre-trained models' weights are frozen and no longer updated.

\paragraph{Feature Extractors.}\label{sec:extractors}
Besides our proposed methodology, we tested and compared the performance of several combination modules. All solutions act as interchangeable modules inside the \texttt{Feature Extractor}, i.e. only need to replace the \textcolor{orange}{yellow} component in Figure \ref{fig:main_architecture}. Feature Extractors' output is a linear representation that is provided as input to the \texttt{Fully-Connected Network}. Figures \ref{fig:com1}-\ref{fig:com2} showcase graphically all the extractors; here, we report a brief rundown across all solutions:
\begin{itemize}
    \item \texttt{Linear} (LIN): pre-trained models' representations are linearized and concatenated.
    \item \texttt{Fixed Linear} (FIX): embeddings are linearized to a predefined size, possibly using adapters to scale the information.
    \item \texttt{Convolutional} (CNN): pre-trained models' outputs are concatenated along the channel dimension. They are processed by a predefined number of convolutional layers, and the resulting information is flattened to linear.
    \item \texttt{Mixed} (MIX): this is a combination of the previous methods. Data coming from different spaces are dealt separately and then combined.
    \item \texttt{Reservoir} (RES): inspired by \citet{gallicchio2017}, this approach leverages reservoir layers to combine models' representations.
    \item \texttt{Dot Product Attention} (DPA): via \texttt{scaled dot product attention} \citep{vaswani2017attention} we compute the final weighted representation using as query vector the representation obtained from the State Encoder.
\end{itemize}

\paragraph{Initial Experiments.}\label{sec:init_exp}
We chose three different Atari games as benchmark, with different levels of difficulty: \texttt{Pong}, \texttt{Ms.Pacman}, and \texttt{Breakout}. After training each pre-trained model, we run a first round of experiments for the three games to select the three best performing extractors and their respective embedding size, which will be used for our empirical analysis along with our proposed method. Agents are trained using PPO \citep{schulman2017proximal} for \texttt{10M} steps using the default parameters. The \texttt{Fully-Connected Network} and the \texttt{Shared Weight Network} are defined as a single layer MLP.
Appendix \ref{sec:app-com-mod} reports all the possible configurations, a detailed analysis on the computational cost and their performance in this initial experimental phase. We report the best performing modules for each game, the chosen embedding size is detailed between parentheses: \texttt{\textbf{Pong}}: WSA (1024), RES (1024), CNN (2); \texttt{\textbf{Ms.Pacman}}: WSA (256), RES (1024), CNN (2); \texttt{\textbf{Breakout}}: WSA (256), FIX (512), CNN (3).

\subsection{Comparison with End-to-End model (\texttt{Q1})}\label{sec:results_1}
\begin{figure}[!h]
    \centering
    \begin{subfigure}[b]{0.32\textwidth}
        \centering
        \includegraphics[width=\textwidth]{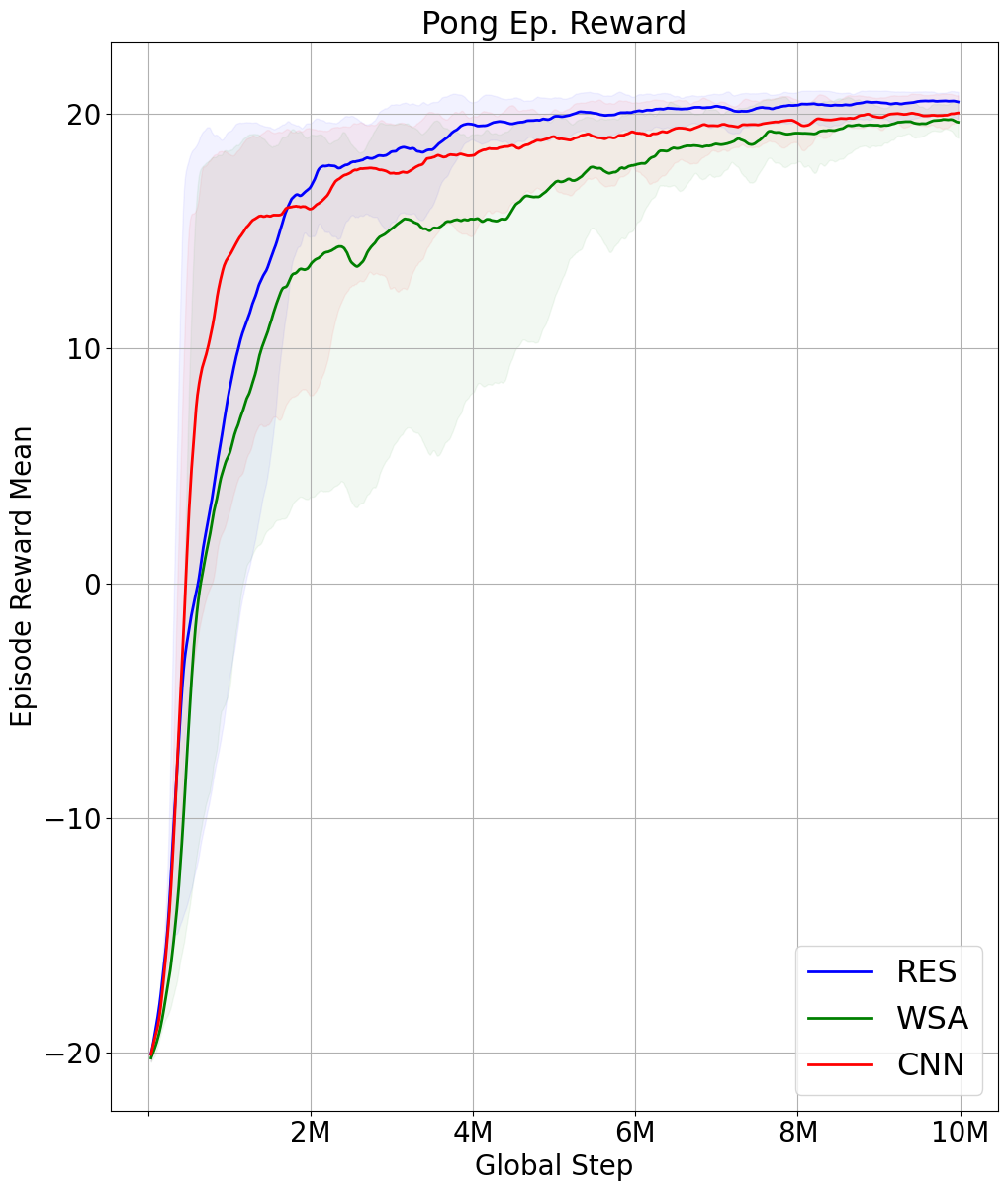}
        \caption{\texttt{Pong}}
        \label{fig:pongtraining}
    \end{subfigure}
    \hfill
    \begin{subfigure}[b]{0.32\textwidth}
        \centering
        \includegraphics[width=\textwidth]{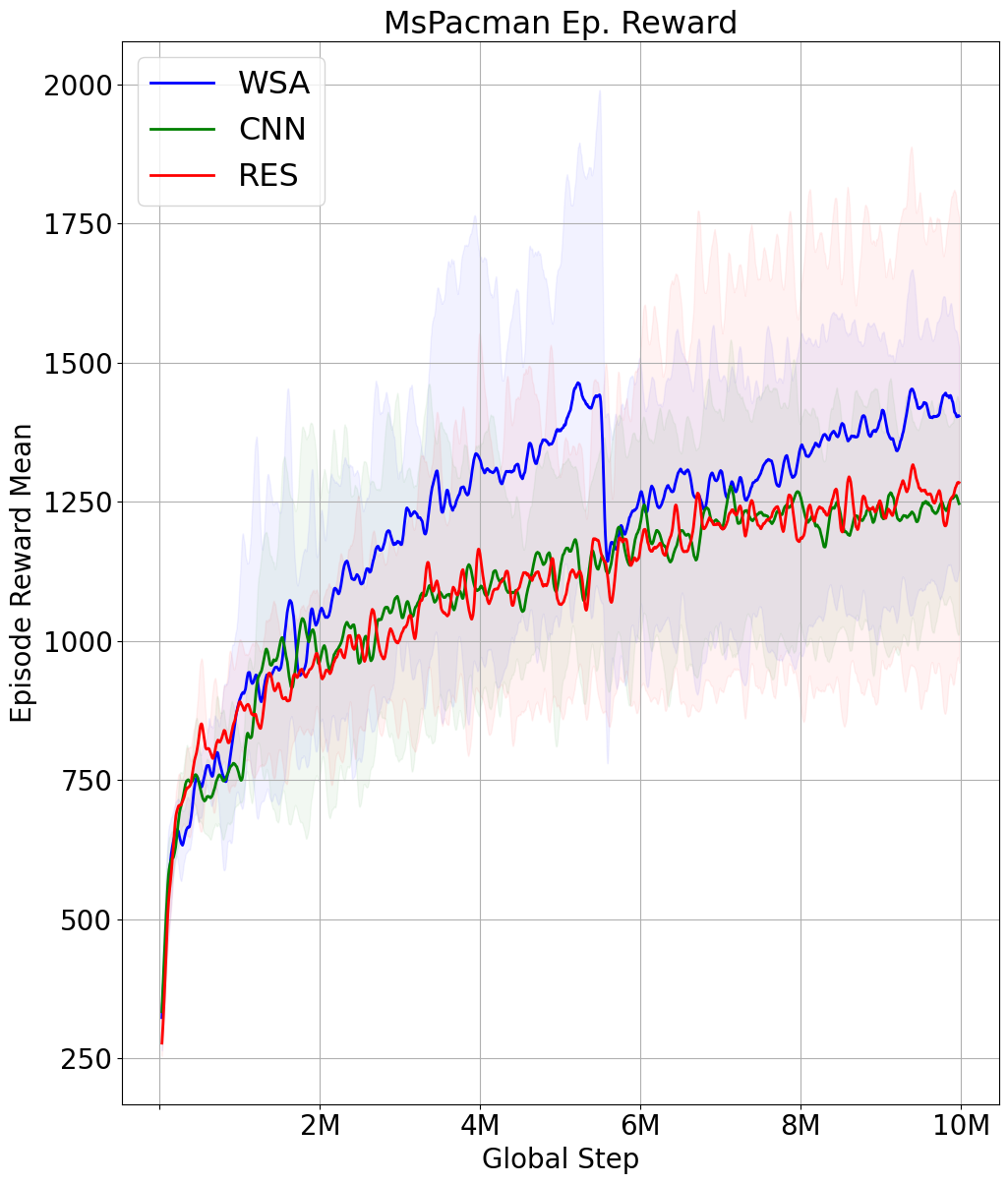}
        \caption{\texttt{Ms.Pacman}}
        \label{fig:mspacmantrain}
    \end{subfigure}
    \hfill
    \begin{subfigure}[b]{0.32\textwidth}
        \centering
        \includegraphics[width=\textwidth]{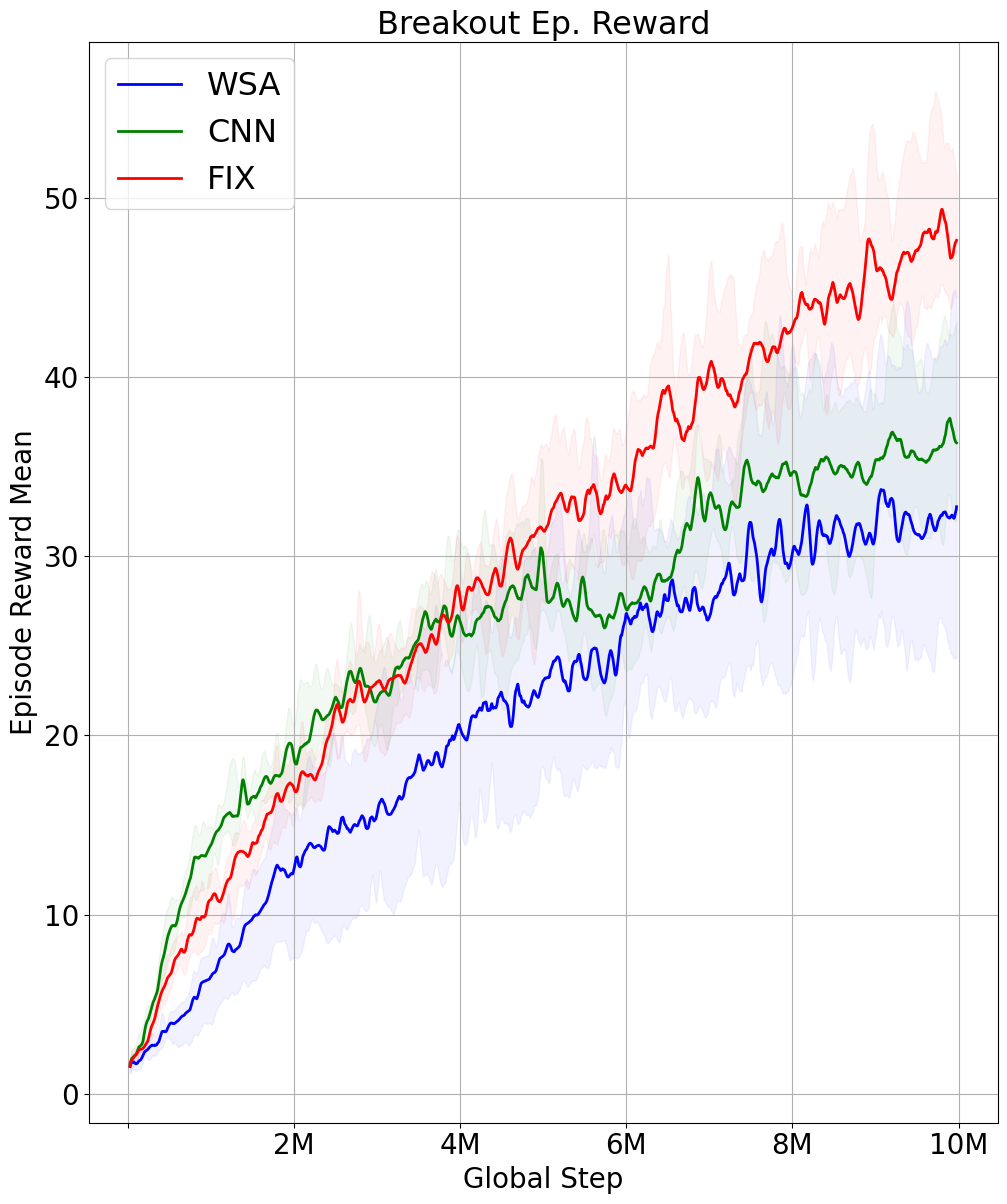}
        \caption{\texttt{Breakout}}
        \label{fig:breakouttrain}
    \end{subfigure}
    \caption{Cumulative reward during \texttt{training} of different agents using WSA and other combination modules on three Atari games. Each subfigure shows the mean score, with shaded areas indicating the standard deviations across multiple agents.}
    \label{fig:trainresults}
\end{figure}

The core idea is to assess if WSA is capable of combining out-of-the-box multiple pre-trained models and to evaluate if the enriched representation can be exploited for learning. It is important to note that the results are obtained without conducting any hyperparameter search. This analysis strictly focuses on the quality and insights of the computed representation.
In the following empirical evaluation, we used the methodology and setup discussed for the initial experiments. For each game, multiple agents instances are trained using different seeds. For a comprehensive view of agents' performance, the score is averaged over different versions, agents are tested across \texttt{5 seeds} (different from the training ones) for \texttt{50 episodes}. Figure \ref{fig:trainresults} shows the average learning curves of agents using different combination modules during training - the shaded area depicts the standard deviation. Table \ref{tab:results} reports the averaged results during evaluation. In this analysis, we compare our architecture with the classical end-to-end model (E2E) used in Atari games. We also include three additional baselines: (\textit{i}) \textit{Choose-1} (C1), that uses only Swin \citep{liu2021swin} - a single large pre-trained model to compute the enriched representation; (\textit{ii}) \textit{Ensamble} (ENS), where the representation is obtained by averaging the pre-trained models' embeddings; (\textit{iii}) \textit{Full Training} (FT), where the models are not pretrained but learn their weights alongside the policy ones.

Focusing on the first two games, Figures \ref{fig:pongtraining}-\ref{fig:mspacmantrain}, WSA and other combination modules yield a high reward in the early stages of training. This suggests that pre-trained models provide an insightful and comprehensive base of knowledge on the environment representation. Looking at the evaluation results in Table \ref{tab:results}, WSA matches the maximum reward (\texttt{21}) on \texttt{Pong} and achieves a higher score (\texttt{2530}) on \texttt{Ms.Pacman} than E2E. We address \texttt{Breakout} results in Section \ref{sec:breakout_study} to answer (\texttt{Q2}). An initial consideration concerns the difference between training and evaluation scores. This is particularly evident in \texttt{Ms.Pacman}, where WSA provides a solid generalization for the task, scoring around \texttt{1400} during training compared to \texttt{2530} during evaluation. All the additional baselines perform worse than our proposed architecture and the end-to-end model in the two more complex games, \texttt{Ms.Pacman} and \texttt{Breakout}; only ENS showcases competitive performance in Pong. A reason for the failure of C1 can be attributed to the out-of-distribution scenario of the Atari games compared to the training data of the Swin model, resulting in vague and incoherent embeddings. The poor performance of the ENS baseline further highlights the need for a smarter and more complex way to combine multiple embeddings.

\begin{table}[ht]
\begin{minipage}[b]{0.5\linewidth}
Simply averaging the embeddings’ values fails to help learning competitive agents’ policies. The full-training baseline shows how the pre-trained models effectively provide insightful representation that cannot otherwise be learned from scratch while optimizing agents’ performance. As a sanity check, to ensure agents' performance does not depend on the particular RL algorithm, we also compare WSA effectiveness using \texttt{DQN}: training curves and evaluation scores are reported in Appendix \ref{sec:dqn-comp}. 

\paragraph{Robotic Manipulation} We include also an additional experiment using \texttt{ManiSkill} \citep{taomaniskill3}, a powerful framework for robot simulation, especially manipulation skills, and training powered by SAPIEN \citep{Xiang_2020_SAPIEN}. In this scenario we consider the \texttt{Push-Cube} environment, where the robot arm has to move a block to a target location. To process the visual representation, we employed different open-source and off-the-shelf pre-trained models. Unlike the \texttt{Atari} environments, this setting involves continuous control and presents a more challenging task. Despite the increased complexity of this setting, with just a small change in the architecture \texttt{WSA} demonstrates strong performance achieving \textbf{90\%} success during evaluation. Furthermore, we exploit this experimental setup to benchmark WSA against two monolithic methods, InstructRL \citep{liu2022instruction} and OpenVLA \citep{kim2024openvla}, which build upon pre-trained multi-modal models and transformers architectures. Despite using only a fraction of the training time, WSA masters the task, whereas the other methods struggle to adapt within the same time and computational budget, highlighting WSA’s efficiency and adaptability to different scenarios. We provide an extended analysis and discussion of these experimental results in Appendix \ref{sec:maniskill}.
\end{minipage}\hfill
\begin{minipage}[b]{0.45\linewidth}
\centering
    \begin{tabular}[b]{lll}
                \multicolumn{1}{l}{Environment}  &\multicolumn{1}{l}{\bf Agent} &\multicolumn{1}{l}{\bf Reward} \\
                \hline \\
                \multirow{6}{*}{\texttt{Pong}}
                                      & C1 & -21 $\pm$ 0.00 \\
                                      & ENS & 20.38 $\pm$ 0.46 \\
                                      & FT & -20.29 $\pm$ 0.25 \\
                                      & CNN & 21 $\pm$ 0.00 \\
                                      & RES & 20.85 $\pm$ 0.29 \\
                                      & WSA & 21 $\pm$ 0.00 \\
                                      & E2E & 20.51 $\pm$ 0.69 \\
                                      \\ \hline \\
                \multirow{6}{*}{\texttt{Ms.Pacman}}
                                      & C1 & 653.595 $\pm$ 47.14 \\
                                      & ENS & 642.87 $\pm$ 58.96 \\
                                      & FT & 689.47 $\pm$ 28.04 \\
                                      & CNN & 1801.30 $\pm$ 20.95 \\
                                      & RES & 1369.27 $\pm$ 565.23 \\
                                      & WSA & 2530.20 $\pm$ 23.09 \\
                                      & E2E & 2145.80 $\pm$ 167.16 \\
                                      \\ \hline \\
                \multirow{7}{*}{\texttt{Breakout}}
                                      & C1 & 8.38 $\pm$ 2.41 \\
                                      & ENS & 62.30 $\pm$ 7.50 \\
                                      & FT & 8.89 $\pm$ 2.04 \\
                                      & CNN & 65.98 $\pm$ 1.62 \\
                                      & FIX & 87.17 $\pm$ 6.87 \\
                                      & WSA & 99.58 $\pm$ 6.66 \\
                                      & WSA \texttt{(M)} & 345.52 $\pm$ 6.47 \\
                                      & E2E & 404.46 $\pm$ 13.49 \\
    \end{tabular}
    \caption{Performance during \texttt{evaluation} averaged across 5 different seeds. E2E model scores are computed using the \texttt{Baselines PPO} results collected in Open RL Benchmark \citep{openrl2024}.}
    \label{tab:results}
\end{minipage}
\end{table}

\subsection{Out of Distribution (Q2)}\label{sec:breakout_study}
\begin{figure}[ht]
    \centering
    \begin{subfigure}[b]{0.45\textwidth}
        \centering
        \includegraphics[width=0.7\textwidth]{images/breakout_train_matplotlib.png}
        \caption{\texttt{Breakout}}
        \label{fig:breakouttrain_2}
    \end{subfigure}
    \hfill
    \begin{subfigure}[b]{0.45\textwidth}
        \centering
        \includegraphics[width=0.7\textwidth]{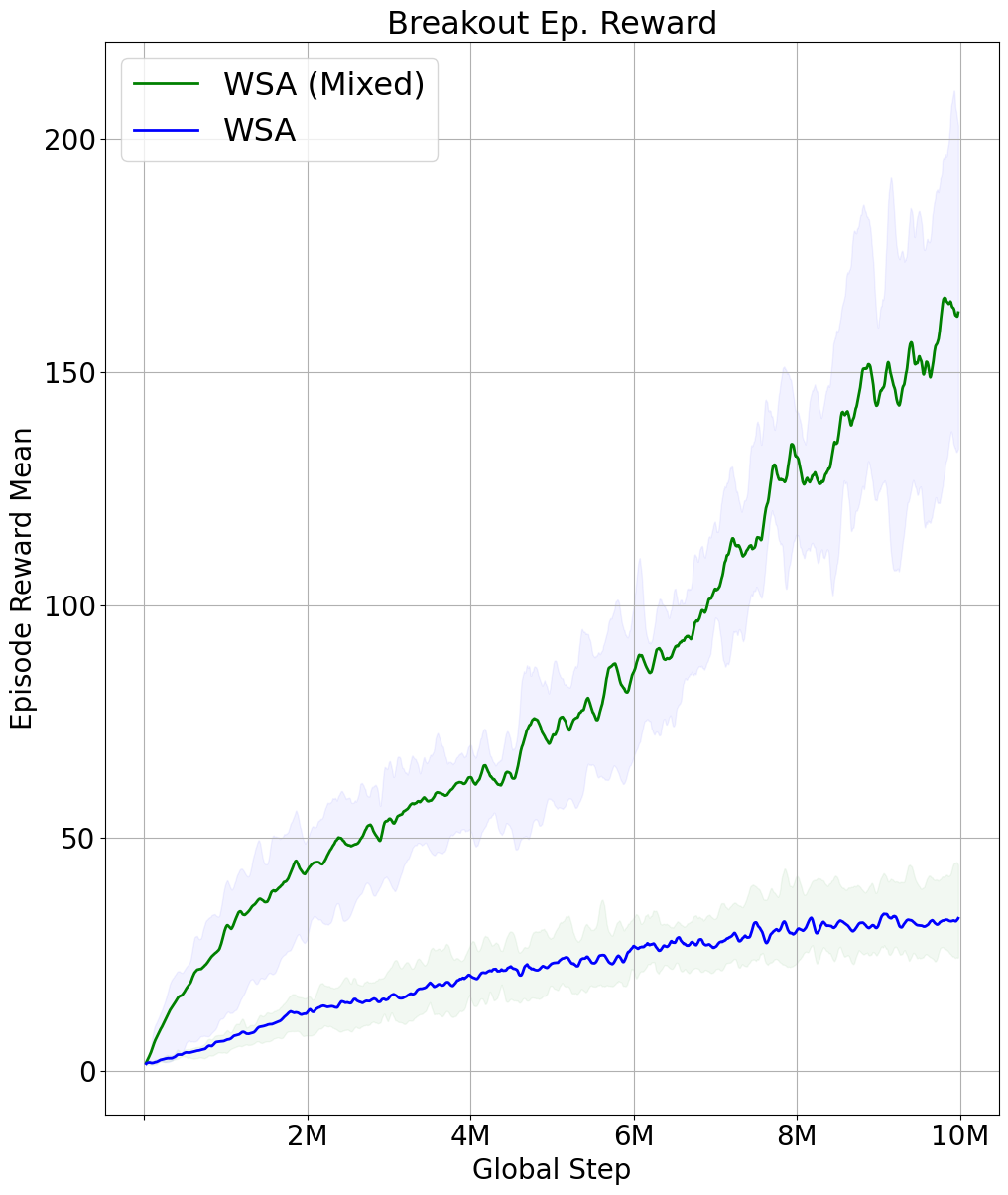}
        \caption{Using Mixed Data}
        \label{fig:breakout_expert}
    \end{subfigure}
    \caption{Performance comparison of WSA across different strategies on \texttt{Breakout}. Each subfigure displays the average score with the standard deviation shaded. They report the different experiments to improve the performance of WSA: (\ref{fig:breakouttrain_2}) shows the initial condition, (\ref{fig:breakout_expert}) pre-training models using random and expert data.}
    \label{fig:breakout_study}
\end{figure}

Unexpectedly on \texttt{Breakout}, WSA did not work straightaway. Both in Figure \ref{fig:breakouttrain} and in Table \ref{tab:results}, the performance of WSA is extremely low. Unlike \texttt{Pong} and \texttt{Ms.Pacman}, where the game screen remains relatively stable, \texttt{Breakout} dynamically evolves as more blocks are removed with score progression. This poses a \textbf{distributional shift} problem between training and test data. Our initial training data, collected from random agents, lacks late-game scenarios with few blocks remaining, which leads to limitations in the model's performance during inference. To tackle this problem and show the consequences of a limited training dataset, we gather new datasets from both random and expert agents to encompass early and late-game scenarios. The problem of updating knowledge in pre-trained models has been studied in several works \citep{cpt, cptcossu}. Moreover, how to continuously update models without occurring in catastrophic forgetting benefits from the plethora of methodologies studied continual learning \citep{surveycl}. We retrain all the pre-trained models and RL agents. Significant improvements are observed, as shown in Figure \ref{fig:breakout_expert} and Table \ref{tab:results}, using \texttt{(M)}. Notably, WSA demonstrates a substantial increase in the agent's final score (\texttt{99 $\rightarrow$ 345}), positioning itself slightly below the E2E model performance, and echoing the trends analyzed in Section \ref{sec:results_1} for the other two Atari games. Additionally, Appendix \ref{sec:breakout-add} reports the same analysis for the other combination modules presented in Figure \ref{fig:breakouttrain_2}.

\subsection{Extended Experiments: optimized WSA}
In Section \ref{sec:results_1}, we analyzed whether WSA could serve as a valid alternative to the convolutional layers that characterize end-to-end agents on Atari. We showed that the information obtained by the combination of multiple pre-trained models enhances the performance of agents. In this section, we aim to evaluate the potential of WSA on a broader set of Atari games by also optimizing the hyperparameters of the RL algorithm. To perform model optimization we leverage the CleanRL implementation of our method to incorporate WSA agents inside PufferLib \citep{pufferlib, suarez2025pufferlib}, which provides a faster vectorization reducing significantly the training time, and use CARBS \citep{carbs} as hyperparameter optimization algorithm. For each game, we run a set of sweeps consisting of 50/100 runs (2-3 days of compute time) to identify the optimal configurations. The search space is reported in Appendix \ref{sec:app-carbs}. We train the best-performing configuration and, in Table \ref{tab:ppo_wsa_comparison}, we report the evaluation performance across multiple seeds. Between parenthesis we include Human Normalized Score (HNS) and Capped-HNS (CNHS) \citep{agent57} of the best agent.\footnote{Data used for computation is taken from \href{https://github.com/vwxyzjn/ppo-atari-metrics/blob/main/hns.py}{\textit{CleanRL-HNS}}.} For the E2E architecture we report the score from \citet{openrl2024}.

\begin{table}[h]
    \centering
    \begin{tabular}{l|cc}
        \bf Game & \bf E2E & \bf WSA \\
        \midrule
        Asteroids & $1511.16 \pm 117.01_{(0.0195/0.0195)}$ & $2130.95 \pm 139.96_{(0.0333/0.0333)}$ \\
        Beam Rider & $2842.55 \pm 423.97_{(0.1753/0.1753)}$ & $1443.85 \pm 233.17_{(0.0793/0.0793)}$ \\
        Breakout & $404.46 \pm 13.49_{(14.4531/1.0)}$ & $395.93 \pm 27.59_{(14.6465/1.0)}$ \\    
        Enduro & $1061.93 \pm 41.05_{(1.2818/1.0)}$ & $1113.23 \pm 69.29_{(1.3742/1.0)}$ \\      
        MsPacman & $2145.80 \pm 167.16_{(0.3019/0.3019)}$ & $2502.55 \pm 141.23_{(0.3517/0.3517)}$ \\
        Pong & $20.51 \pm 0.69_{(1.1810/1.0)}$ & $20.62 \pm 0.37_{(1.1810/1.0)}$ \\              
        Qbert & $15298.00 \pm 1011.65_{(1.2148/1.0)}$ & $9194.75 \pm 503.5_{(0.7173/0.7173)}$ \\ 
        Seaquest & $1517.73 \pm 399.81_{(0.0440/0.0440)}$ & $2795.31 \pm 288.35_{(0.0718/0.0718)}$ \\
        Space Invaders & $1038.64 \pm 72.93_{(0.6336/0.6336)}$ & $833.94 \pm 65.06_{(0.4939/0.4939)}$ \\
        \midrule
        \textit{Mean HNS/CHNS} & $2.1450 / 0.5749$ & $2.1054 / 0.5275$ \\
    \end{tabular}
    \caption{Performance comparison of PPO and WSA on various Atari games, between parenthesis are reported HNS and CHNS, respectively. E2E model scores are computed using the \texttt{OpenAI/Baselines PPO} results collected in Open RL Benchmark \citep{openrl2024}.}
    \label{tab:ppo_wsa_comparison}
\end{table}

Out of the nine games considered in this evaluation, WSA outperforms E2E in five of them (\texttt{Asteroids}, \texttt{Enduro}, \texttt{MsPacman}, \texttt{Pong}, \texttt{Seaquest}) and shows comparable performance in \texttt{Breakout}, as indicated by overlapping confidence intervals. On the other hand, in three games (\texttt{Beam Rider}, \texttt{Qbert}, \texttt{Space Invaders}) WSA performs worse then E2E. The small difference in the averaged human normalized scores, less than 0.05 in the capped case, suggests that WSA performs in the same range as the baseline, while providing benefits such as modularity, interpretability (Section \ref{sec:scaling}), and generalization advantages (Section \ref{sec:rob}, Appendix \ref{sec:app-hackatari}). Given the characteristics of the games, we hypothesize that the performance is influenced by a similar problem to the one analyzed in section \ref{sec:breakout_study}. For example, in later stages of \texttt{Space Invaders}, aliens are removed from the screen and the structures are damaged, picturing a different scenario compared to the initial conditions, leading to poor performance of the pre-trained models.

\subsection{WSA Robustness (Q3)}\label{sec:rob}
Differently from end-to-end approaches, WSA leverages the embeddings provided by the pre-trained models, which are trained via supervised or unsupervised learning on a set of collected data and are kept frozen during the RL learning phase. For this reason, we expect the representations learned by these models to be more robust with respect to convolutional layers that are learned together with the policy (which tend to over-specialize to the training environment). To investigate this issue and measure the positive impact of WSA, we use \texttt{HackAtari} \citep{hackatari}, which provides different variations of Atari gaming by changing game elements' color or altering the game settings. In our analysis, we consider two games to evaluate WSA in two different scenarios. The former is \texttt{Pong}, in its variation \textit{Lazy Enemy} (LE) where the CPU behavior is altered allowing movement only after the player hits the ball. The latter is \texttt{Breakout}, where we use multiple variations that change the \textit{color} either of the \textit{player} (CP) or \textit{blocks} (CB) (both can vary among five possible variation corresponding to the colors black, white, red, blue and green). In Table \ref{tab:hackatari_table}, we report the results of the evaluation on the two settings, averaged over 30 episodes. For \texttt{Breakout}, since there are no results in the original work, we train and evaluate both E2E and WSA agents. In Appendix \ref{sec:app-hackatari}, we report the training curves of the agents, as well as a complete table showcasing the performance on all 25 possible variations of Breakout. WSA is more robust and resilient to environmental changes compared to E2E in both scenarios. In \texttt{Pong}, WSA maintains a \textbf{winning} score (positive), while in \texttt{Breakout}, it performs more than \textbf{3x better} than E2E in all configurations.

\begin{table}[h]
    \centering
    \begin{tabular}{c|c|c|c}
        Game & Random & E2E & WSA\\
        \addlinespace
        \hline 
        \addlinespace
         \begin{tabular}{cc} 
            \textbf{Training} \\
            \textbf{Testing}
         \end{tabular} &
         \begin{tabular}{cc} 
            - \\
            variation
        \end{tabular} &
        \begin{tabular}{cc} 
            original \\
            variation
        \end{tabular} &
        \begin{tabular}{cc}
            original \\
            variation
        \end{tabular} \\
        \addlinespace
        \hline
        \addlinespace
        Pong (LE) & $-20.1 \pm 0.4$ & $-12.6 \pm 2.4$ & $13.56 \pm 6.43$ \\
        \addlinespace
        \hline
        \addlinespace
        Breakout (CP 0) (CB \texttt{*}) & $1.19 \pm 1.23$ & $12.14 \pm 7.50$ & $45.64 \pm 18.88$ \\
        Breakout (CP 1) (CB \texttt{*}) & $1.18 \pm 1.17$ & $13.54 \pm 9.37$ & $45.35 \pm 17.22$ \\
        Breakout (CP 2) (CB \texttt{*}) & $1.31 \pm 1.24$ & $13.07 \pm 9.06$ & $44.04 \pm 16.61$ \\
        Breakout (CP 3) (CB \texttt{*}) & $1.00 \pm 1.07$ & $13.10 \pm 8.82$ & $43.49 \pm 16.30$ \\
        Breakout (CP 4) (CB \texttt{*}) & $1.16 \pm 1.16$ & $12.41 \pm 7.96$ & $44.70 \pm 17.84$ \\
        \addlinespace

        Breakout (CP \texttt{*}) (CB 0) & $1.18 \pm 1.15$ & $12.81 \pm 7.78$ & $44.93 \pm 16.95$ \\
        Breakout (CP \texttt{*}) (CB 1) & $1.15 \pm 1.16$ & $13.31 \pm 9.54$ & $45.20 \pm 16.92$ \\
        Breakout (CP \texttt{*}) (CB 2) & $1.24 \pm 1.23$ & $12.49 \pm 8.15$ & $44.48 \pm 17.15$ \\
        Breakout (CP \texttt{*}) (CB 3) & $1.07 \pm 1.20$ & $12.83 \pm 8.19$ & $44.75 \pm 19.06$ \\
        Breakout (CP \texttt{*}) (CB 4) & $1.20 \pm 1.14$ & $12.82 \pm 9.07$ & $43.86 \pm 16.78$ \\
    \end{tabular}
    \caption{Results averaged over multiple episodes for different configurations of the game as in \citet{hackatari}. The table presents performance for random, PPO, and WSA agents across variations of Pong and Breakout. Scores marked with \texttt{*} indicate averages over all possible combinations for that particular setting.}
    \label{tab:hackatari_table}
\end{table}

\subsection{Scaling Number of Models (Q4)}\label{sec:scaling}
\begin{figure}[h]
    \centering
    \begin{subfigure}[b]{0.49\textwidth}
        \centering
        \includegraphics[width=\textwidth]{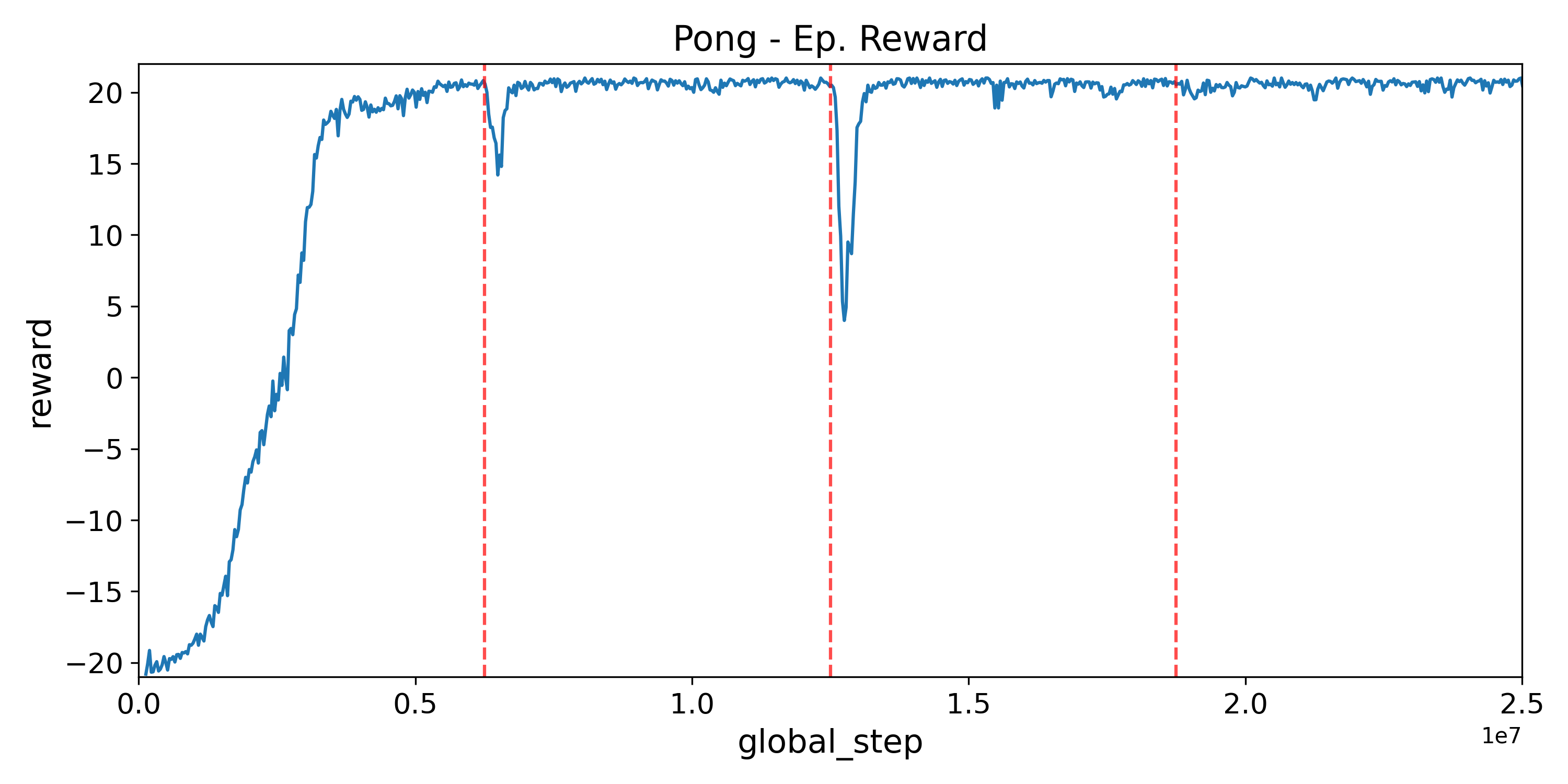}
        \caption{Adding Scenario}
        \label{fig:wsa_add}
    \end{subfigure}
    \hfill
    \begin{subfigure}[b]{0.49\textwidth}
        \centering
        \includegraphics[width=\textwidth]{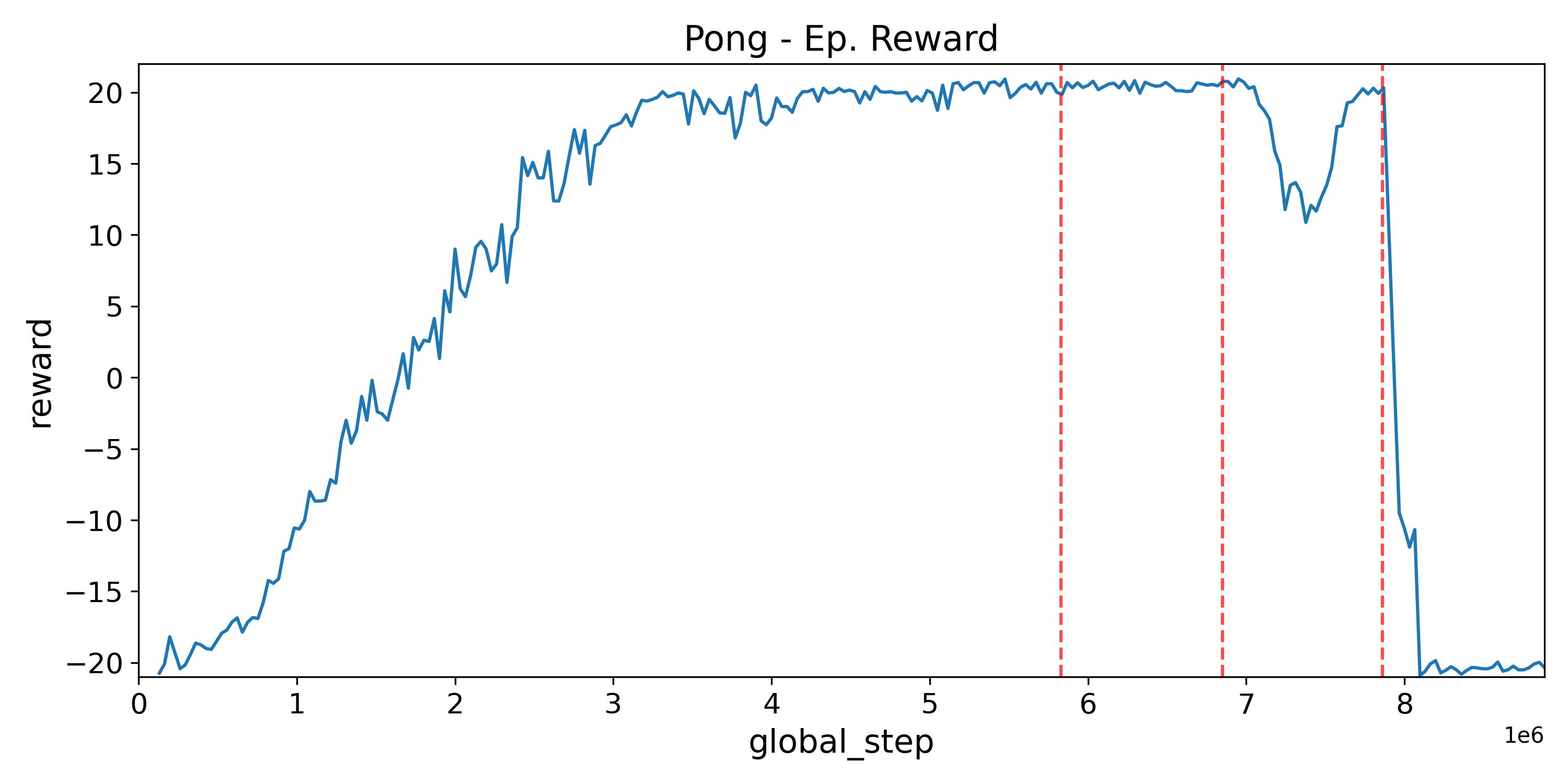}
        \caption{Removal Scenario}
        \label{fig:wsa:remove}
    \end{subfigure}
    \caption{Figures illustrate the adding and removing experiments, red lines mark when a model is added or removed.}
    \label{fig:model_evol}
\end{figure}

\begin{figure}[h]
    \centering
    \begin{subfigure}[b]{0.49\textwidth}
        \centering
        \includegraphics[width=0.8\textwidth]{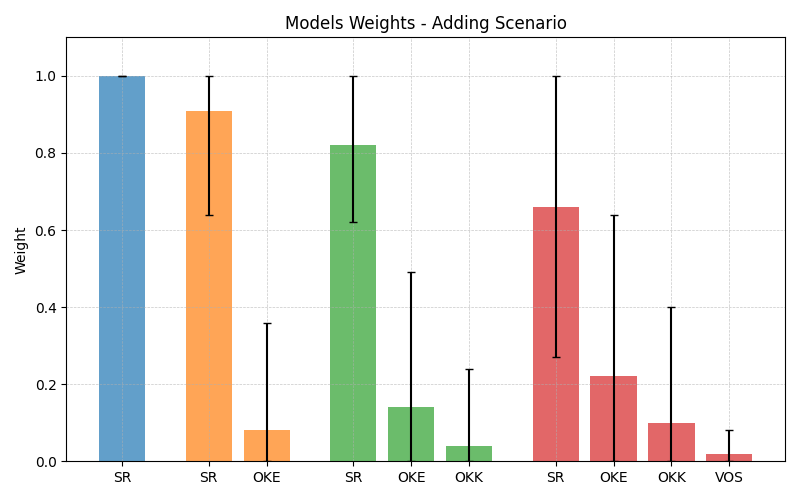}
        \caption{Adding Scenario}
        \label{fig:wsa_add2}
    \end{subfigure}
    \hfill
    \begin{subfigure}[b]{0.49\textwidth}
        \centering
        \includegraphics[width=0.8\textwidth]{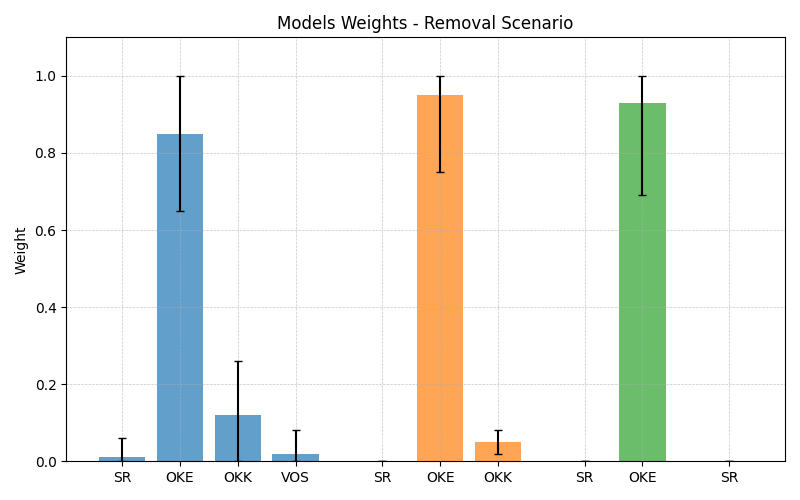}
        \caption{Removal Scenario}
        \label{fig:wsa_remove2}
    \end{subfigure}
    \caption{Average and standard deviation of the weights assigned by our model to each component, represented as bar plots with error bars. Each group corresponds to an evaluation phase, aligned with the red vertical lines in Figure \ref{fig:model_evol}. (\ref{fig:wsa_add2}) shows the Adding Scenario, while (\ref{fig:wsa_remove2}) depicts the Removal Scenario.}
    \label{fig:model_evol2}
\end{figure}

Our approach is designed to ensure two key properties: WSA is both \texttt{scalable} and \texttt{adaptable} to any number of pre-trained models. The \texttt{Shared Weight Network} is capable of managing an arbitrary number of models, making it a versatile solution for dynamically evolving agents. This section examines two distinct scenarios in which the number of pre-trained models changes over time, either increasing or decreasing.
In the \textit{adding} experiment, Figures \ref{fig:wsa_add}, a new model becomes available at regular intervals, marked in red, and it is incorporated into the computation of the embedding while the agent continues training. This setup simulates a scenario where an agent must continuously integrate new knowledge without disrupting its learning process.
On the other hand, in the \textit{removal} experiment, Figure \ref{fig:wsa:remove}, the agent is initially trained with the \textit{complete set} of available models, allowing it to achieve peak performance. Subsequently, we systematically remove models from the pool at regular intervals, shown in red. This experiment evaluates the agent’s ability to retain and generalize knowledge as resources diminish, assessing its adaptability to a progressively reduced set of pre-trained models while maintaining performance.
These experiments provide insight into WSA’s flexibility in managing dynamic model availability.In both scenarios, the pre-trained models are ordered as follows: (1) State-Representation (SR), (2) Object Key-Points Encoder (OKE), (3) Object Key-Points KeyNet (OKK), and (4) Video Object Segmentation (VOS). In the adding experiments, pre-trained models are added following the order 1 $\rightarrow$ 4, while in the removal models are removed following the order 4 $\rightarrow$ 1, as also suggested by the labels in Figures \ref{fig:wsa_add2}-\ref{fig:wsa_remove2}. Looking at Figure \ref{fig:wsa_add2}, as new models are added to the pool, WSA slowly adapts its weights to leverage embeddings from all the models while maintaining optimal performance. In the removal scenario, Figure \ref{fig:wsa_remove2}, at the end of the training, WSA assigns the highest importance to the OKE model. Removing other models does not affect performance and WSA is able to adapt, once this model is removed, the performance collapses because the remaining model does not provide useful information.

\begin{figure}[ht]
    \begin{center}
        \includegraphics[width=0.8\textwidth]{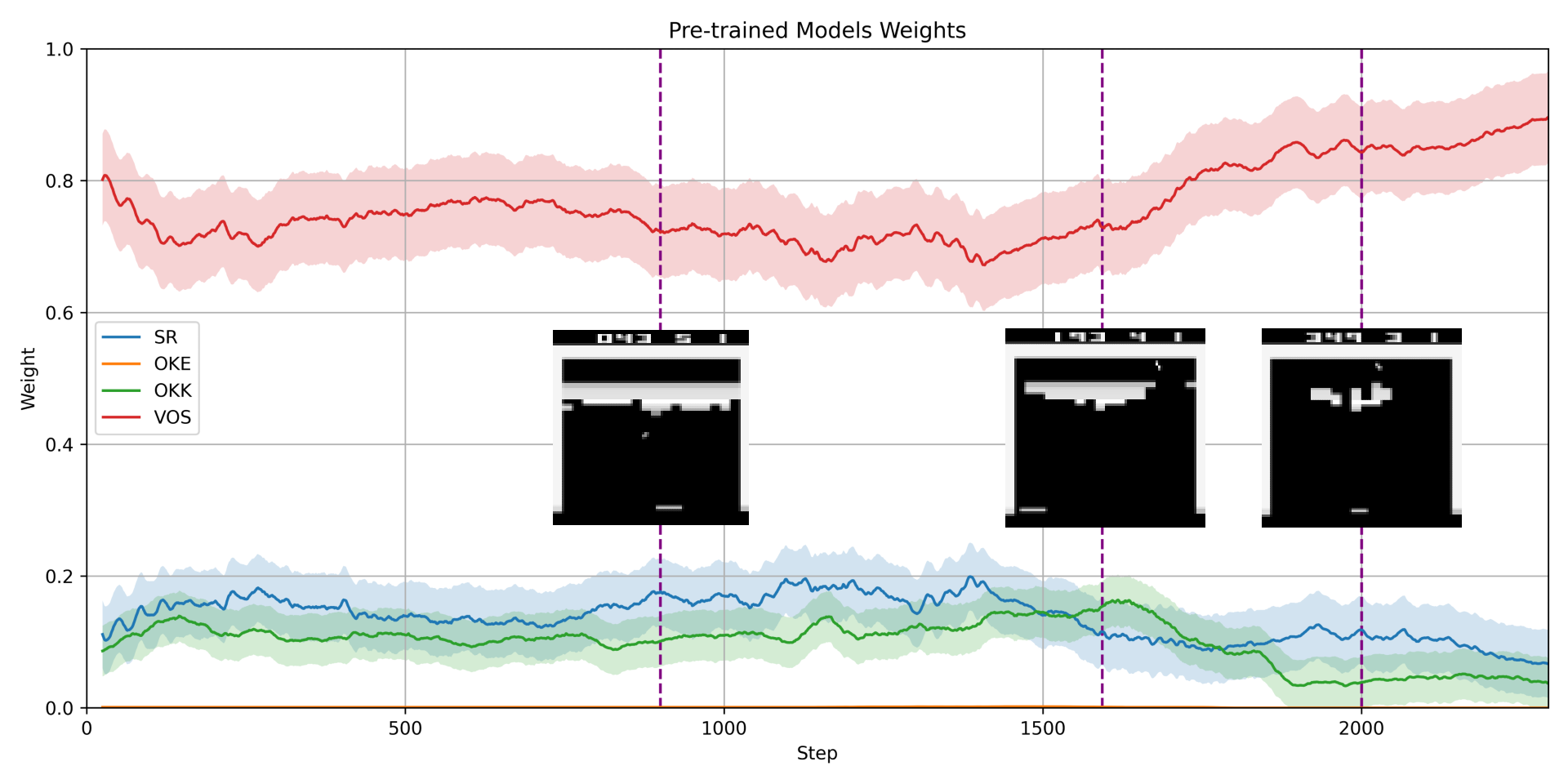}
    \end{center}
    \caption{Weights assigned by WSA to different pre-trained models during rollouts in the Breakout environment. The embedded game frames highlight representative gameplay situations. In the early stages, WSA distributes attention across multiple models, whereas in the later stages, it converges towards relying predominantly on a single model, indicating adaptive specialization based on the game scenario.}
    \label{fig:modelw_exp}
\end{figure}

\paragraph{WSA Explainability.}\label{sec:explain}
Lastly, Figure \ref{fig:modelw_exp} illustrates the \texttt{explainability} feature provided by WSA. By examining the current frames and the corresponding weights allocated by the shared network to each pre-trained model, one can appreciate the agents' decision-making process. Moreover, this information helps in debugging and understanding agents behavior in dynamic environments, as the one studied in Figure \ref{fig:model_evol2}. The visualization reveals how various models are leveraged in distinct contexts, shedding light on agents' dynamic adaptation of prior knowledge. In particular, Figure \ref{fig:modelw_exp} shows how WSA's weights change throughout an episode on \texttt{Breakout}. WSA assigns the highest weight to the Video Object Segmentation (VOS) model \citep{goel2018unsupervised} and combines it with contributions from other two models \citep{anand2019unsupervised, kulkarni2019unsupervised}. Moreover, we can appreciate how, in the initial stages of the games where the majority of the blocks are still present, WSA combines multiple embeddings. Once the agent gets access to the upper portion of the screen, it starts eliminating most of the blocks and WSA gradually increases the importance of the VOS model. This behavior may be attributed to VOS embeddings that are pre-trained to generate segmentation masks across multiple frames. During this phase of the game, the agent must wait for the ball to hit the blocks and fall, making the ability to track objects over time particularly important.

\section{Conclusion}\label{sec:conclusions}
In this work, we explored how prior knowledge encoded in multiple pre-trained models can be integrated as inductive biases to guide representation learning in Reinforcement Learning agents. We addressed the challenge of combining diverse latent representations into a unified feature space and proposed the Weight Sharing Attention (WSA) module, which uses attention mechanisms and parameter sharing to dynamically balance the contributions of each model in a scalable and model-agnostic way.
Our evaluation across a range of Atari games demonstrates that WSA match or outperform traditional end-to-end architectures, effectively leveraging prior knowledge from multiple sources. These results highlight the potential of combining multiple inductive priors to improve efficiency and policy robustness, particularly in dynamic or changing environments. Looking ahead, we plan to extend our analysis to more diverse and complex scenarios. We also aim to scale our approach by incorporating richer forms of prior knowledge, e.g., Large Language Models (LLMs) or Vision-Language Models (VLMs).

\subsubsection*{Acknowledgments}
We would like to thank the reviewers and Levi H. S. Lelis for their valuable comments. Research partly funded by PNRR - M4C2 - Investimento 1.3, Partenariato Esteso PE00000013 - "FAIR - Future Artificial Intelligence Research" - Spoke 1 "Human-centered AI", funded by the European Commission under the NextGeneration EU programme.

\bibliography{collas2025_conference}
\bibliographystyle{collas2025_conference}

\newpage
\appendix

\section{Experimental Setup}\label{sec:app-exp-setup}
To simulate and interact with the Atari environments \citep{bellemare2013atari} we leverage the API provided by Gymnasium \citep{towers_gymnasium_2023}. The training performance was tracked and recorded using \texttt{Weights and Biases} \citep{wandb}. We build our work\footnote{Code is available at \texttt{https://github.com/EliaPiccoli/Collas-WSA}} using \texttt{Stable-Baselines3} (SB3) \citep{stable-baselines3} and CleanRL \citep{cleanrl} as backbones to create agents' architecture and use reliable implementations of RL algorithms. In particular, we exploit the unconstrained definition of the agent model to decompose the agents' network in \texttt{Feature Extractor} and \texttt{Fully-Connected Network}, as it also highlighted in Figure \ref{fig:main_architecture}. We use the \texttt{Feature Extractor} to define the different combination modules that provide as output the combined latent representation. The \texttt{Fully-Connected Network} receives as input the pre-trained models' unified encoding and maps it to actions (or values). We keep the latter component as simple as possible since most of the information should be provided by the enriched representations. Throughout all the initial experiments and the first comparison analysis, we use the default hyperparameters for different RL algorithms, i.e. PPO and DQN \citep{rl-zoo3}. We do not conduct any hyperparameter search for our agents, unless otherwise stated. Instead, we only modify the model architecture. Even though this setup could be sub-optimal for our solution, we believe is an important test to evaluate the actual influence of combining pre-trained models' representations. Below we report the hyperparamters used for different RL algorithms in the experiments: \texttt{PPO} (Table \ref{tab:ppo_hyperparams}) and \texttt{DQN} (Table \ref{tab:dqn_hyperparams}). These values are the default ones and are also available on \href{https://github.com/DLR-RM/rl-baselines3-zoo}{rl-zoo} \citep{rl-zoo3}. The set of seeds used during evaluation is the following: 47695, 32558, 94088, 71782 and 66638.

\begin{table}[htbp]
    \begin{minipage}{.5\linewidth}
    \centering
        \begin{tabular}{ll}
            \multicolumn{1}{l}{\bf Hyperparameter}  &\multicolumn{1}{l}{\bf Value}
            \\ \hline \\
            N. Envs.       & 8 \\ 
            N. Stacks      & 4 \\
            N. Steps       & 128 \\
            N. Epochs      & 4 \\
            Batch Size     & 256 \\
            N. Timesteps   & 10.000.000 \\
            Learning Rate  & 2.5e-4 \\
            Clip Range     & 0.1 \\
            VF. Coef.      & 0.5 \\
            Ent. Coef.     & 0.01 \\
            Normalize      & True \\
        \end{tabular}
        \caption{PPO hyperparameters.} \label{tab:ppo_hyperparams}
    \end{minipage}%
    \begin{minipage}{.5\linewidth}
        \centering
        \begin{tabular}{ll}
            \multicolumn{1}{l}{\bf Hyperparameter}  &\multicolumn{1}{l}{\bf Value}
            \\ \hline \\
            N. Envs.       & 8 \\ 
            N. Stacks      & 4 \\
            Buffer Size    & 100.000 \\
            Target Update Interval      & 1.000 \\
            Batch Size     & 32 \\
            N. Timesteps   & 10.000.000 \\
            Learning Rate  & 1e-4 \\
            Learning Starts  & 100.000 \\
            Train Freq.    & 4 \\
            Gradient Steps & 1 \\
            Exploration Fraction     & 0.1 \\
            Exploration Final Eps      & 0.01 \\
            Optimize Memory Usage  & False \\
        \end{tabular}
        \caption{DQN hyperparameters.} \label{tab:dqn_hyperparams}
    \end{minipage}
\end{table}

\section{Pre-Trained Models}\label{sec:app-models}
In this section, we provide additional details on the models leveraged as prior knowledge by agents. Tables \ref{tab:obj_keypoints_hyperparams}, \ref{tab:vid_seg_hp}, \ref{tab:state_hp} report the hyperparameters used to train all models; the architecture presented in the original works is kept as the backbone of the models. We re-implement all models and training scripts in PyTorch, which will be released together with our codebase and checkpoints of the models to replicate all the experiments. \texttt{State Representation}'s RAM information is not available for all games; thus, in our experiments, \texttt{Beam Rider} and \texttt{Enduro} have one less pre-trained model due to the unavailabiliy of the information in \citet{anand2019unsupervised}.\\ Next we provide a brief description of each model and the shape of their latent representation.

\begin{itemize}
    \item \texttt{State Representation} \citep{anand2019unsupervised}. Given the current state, it computes a linear representation exploiting the spatial-temporal nature of visual observations.\\
    \textit{Embedding shape}: $[512]$.
    \item \texttt{Object Keypoints} \citep{kulkarni2019unsupervised}. The model is composed by two heads that are used during inference: a convolutional network to compute spatial feature maps and a KeyNet \citep{jakab2018keynet} to predict the keypoint coordinates.\\
    \textit{Embedding shape}: CNN $[ 128 \times 21 \times 21 ]$, KeyNet $[ 4 \times 21 \times 21 ]$.
    \item \texttt{Video Object Segmentation} \citep{goel2018unsupervised}. This model takes as input the last two frames of the game and computes $K$ feature maps that highlights the moving objects. \\
    \textit{Embedding shape}: $[ 20 \times 84 \times 84 ]$.
    \item \texttt{Deep Autoencoder}. It is used to encode the current state and is used in WSA to compute the context. Its architecture is inspired by NatureCNN \citep{mnih2015human}.\\
    \textit{Embedding shape}: $[ 64 \times 16 \times 16]$.
\end{itemize}

\begin{table}[htbp]
  \centering
  \begin{subtable}[t]{.32\linewidth}
    \centering
    \begin{tabular}{ll}
      \multicolumn{1}{l}{\bf Hyperparameters}  &\multicolumn{1}{l}{\bf Value} \\ 
      \hline\\
      Episodes         & 1000  \\
      Max Iter         & 1e6   \\
      Batch Size       & 64    \\
      Image Channels   & 1     \\
      K (n. keypoints) & 4     \\
      Learning Rate    & 1e-3  \\
      LR Decay         & 0.95  \\
      Decay Length     & 1e5   \\
    \end{tabular}
    \caption{}
    \label{tab:obj_keypoints_hyperparams}
  \end{subtable}\hfill
  \begin{subtable}[t]{.32\linewidth}
    \centering
    \begin{tabular}{ll}
      \multicolumn{1}{l}{\bf Hyperparameters}  &\multicolumn{1}{l}{\bf Value} \\ 
      \hline\\
      Num. Frames      & 2      \\ 
      Steps            & 500{,}000  \\
      Batch Size       & 64     \\
      Learning Rate    & 1e-4   \\
      Max Grad. Norm   & 5.0    \\
      Decay Length     & 1e5    \\
      Optimizer        & Adam   \\
      K (n. masks)     & 20     \\
    \end{tabular}
    \caption{}
    \label{tab:vid_seg_hp}
  \end{subtable}\hfill
  \begin{subtable}[t]{.32\linewidth}
    \centering
    \begin{tabular}{ll}
      \multicolumn{1}{l}{\bf Hyperparameters}     &\multicolumn{1}{l}{\bf Value} \\ 
      \hline\\
      Image Width         & 160     \\
      Image Height        & 210     \\
      Grayscaling         & Yes     \\
      Action Repetitions  & 4       \\
      Max-pool (last N)   & 2       \\
      Frame Stacking      & 4       \\
      Batch Size          & 64      \\
      LR (Training)       & 5e-4    \\
      LR (Probing)        & 3e-4    \\
      Entropy Threshold   & 0.6     \\
      Encoder steps       & 80{,}000 \\
      Probe train steps   & 30{,}000 \\
      Probe test steps    & 10{,}000 \\
      Epochs              & 100     \\
      Feature Size        & 512     \\
      Pretraining steps   & 100{,}000 \\
      Num Parallel Envs.  & 8       \\
    \end{tabular}
    \caption{}
    \label{tab:state_hp}
  \end{subtable}
  \caption{Hyperparameter used for three unsupervised pre-trained models, (\ref{tab:obj_keypoints_hyperparams}) Object Keypoints \citep{kulkarni2019unsupervised}, (\ref{tab:vid_seg_hp}) Video Object Segmentation \citep{goel2018unsupervised}, (\ref{tab:state_hp}) State Representation \citep{anand2019unsupervised}.}
  \label{tab:merged_hyperparams}
\end{table}

\section{Additional Experiments}\label{sec:app-add-exp}
\subsection{DQN Comparison}\label{sec:dqn-comp}
\begin{figure}[htbp]
    \centering
    \begin{subfigure}[b]{0.45\textwidth}
        \centering
        \includegraphics[width=0.8\textwidth]{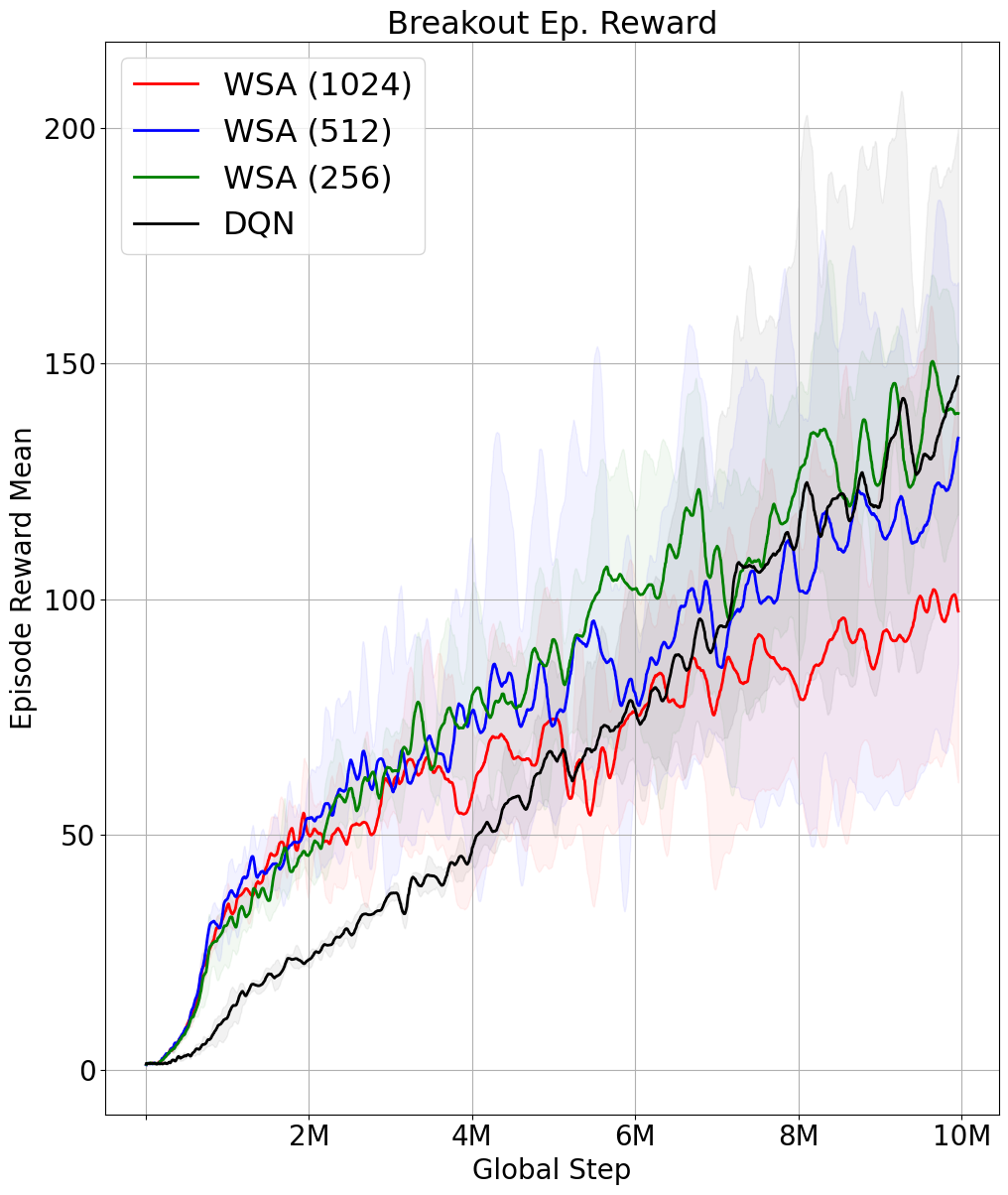}
        \caption{\texttt{Breakout}}
        \label{fig:breakout_dqn}
    \end{subfigure}
    \hfill
    \begin{subfigure}[b]{0.45\textwidth}
        \centering
        \includegraphics[width=0.8\textwidth]{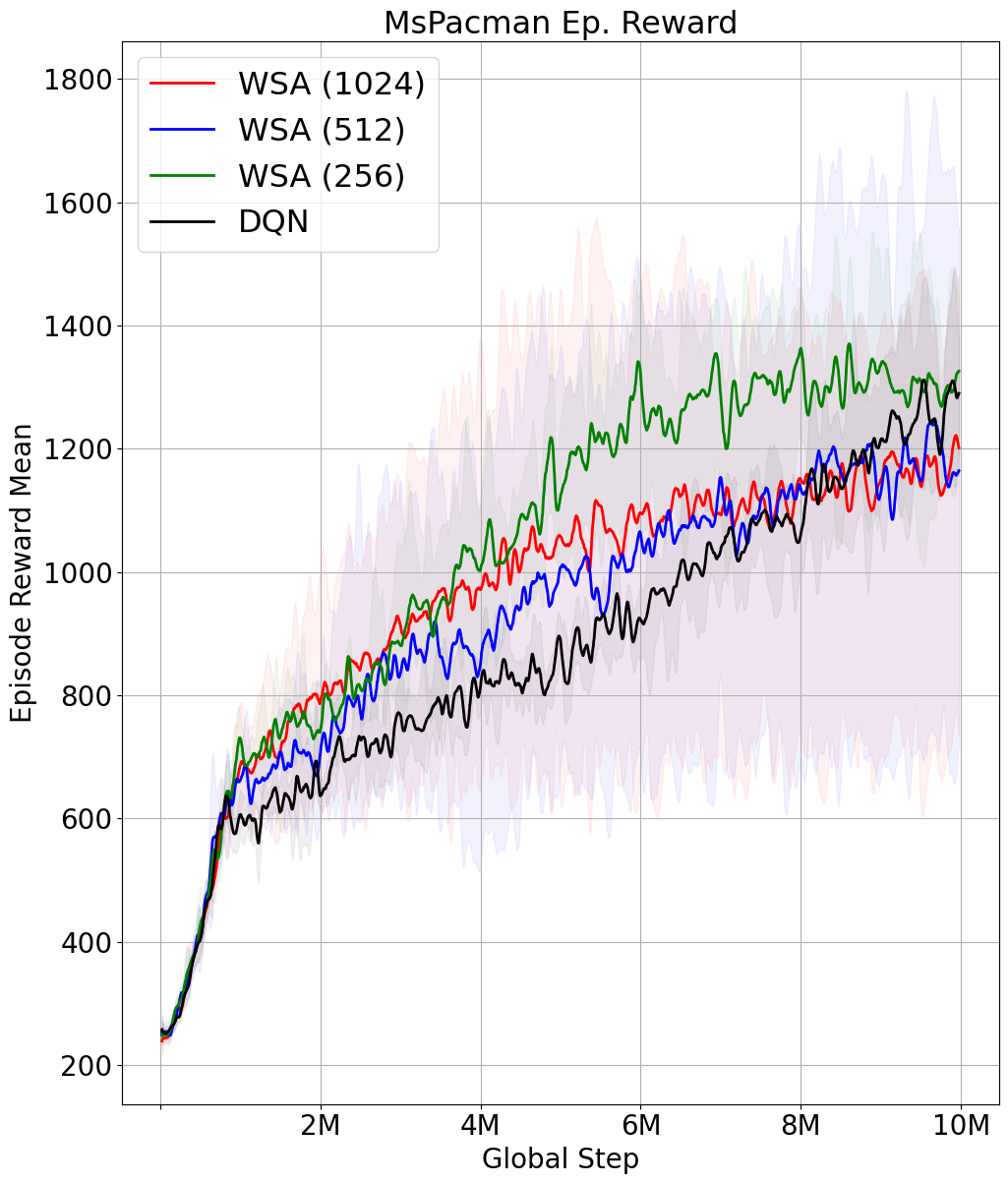}
        \caption{\texttt{MsPacman}}
        \label{fig:mspacman_dqn}
    \end{subfigure}
    \caption{Cumulative reward during \texttt{training} of different agents, comparing WSA and end-to-end model using DQN. Each subfigure shows the mean score, shaded areas indicate the standard deviations across multiple agents.}
    \label{fig:dqn_experiments}
\end{figure}

Using the same methodology and structure of experiments of the main work, we study the performance of WSA while using a different RL algorithm, i.e. \texttt{DQN}. This analysis proves that WSA works well regardless of the algorithm used. In particular, in \texttt{Pong} WSA achieves a perfect score, it \textit{outperforms} the end-to-end architecture in \texttt{Breakout} and \texttt{MsPacman} and in \texttt{SpaceInvaders} WSA achieves a \textit{comparable mean performance} to E2E but with significantly lower variability. Figure \ref{fig:dqn_experiments} depicts the average training trends across different instances with various latent‐representation sizes, while Table \ref{tab:dqn_results} shows the evaluation performance of the best configuration.

\begin{table}[ht]
  \centering
  \begin{tabular}{l|cccc}
    & \bf Breakout & \bf MsPacman & \bf Pong & \bf Space Invaders \\
    \addlinespace
    \hline 
    \addlinespace
    \texttt{WSA} & 213.14 $\pm$ 39.37 & 2047.27 $\pm$ 231.18 & 21 $\pm$ 0.00 & 649.00 $\pm$ 136.85 \\
    & & & & \\
    \texttt{E2E} & 166.65 $\pm$ 20.19 & 1701.00 $\pm$ 490.41 & 20.70 $\pm$ 0.46 & 680.00 $\pm$ 231.26 \\
    \addlinespace
  \end{tabular}
  \caption{Performance during \texttt{evaluation} averaged across 5 different seeds, WSA embedding size is 256.}
  \label{tab:dqn_results}
\end{table}

\pagebreak

\subsection{ManiSkill additional experiments}\label{sec:maniskill}
\begin{figure}[ht]
    \centering
    \begin{subfigure}[b]{0.40\textwidth}
        \includegraphics[width=\textwidth]{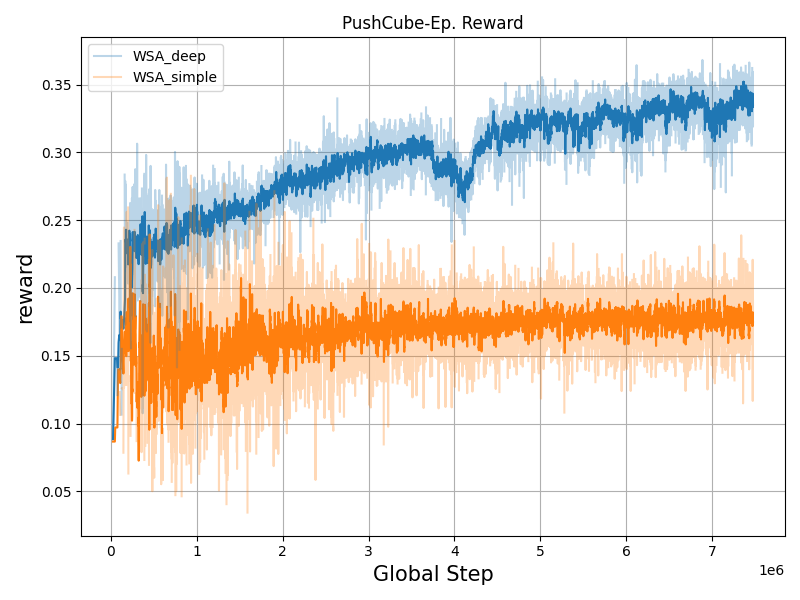}
        \caption{WSA agent mean reward per episode}
        \label{fig:pushcube rew}
    \end{subfigure}
    \begin{subfigure}[b]{0.40\textwidth}
        \includegraphics[width=\textwidth]{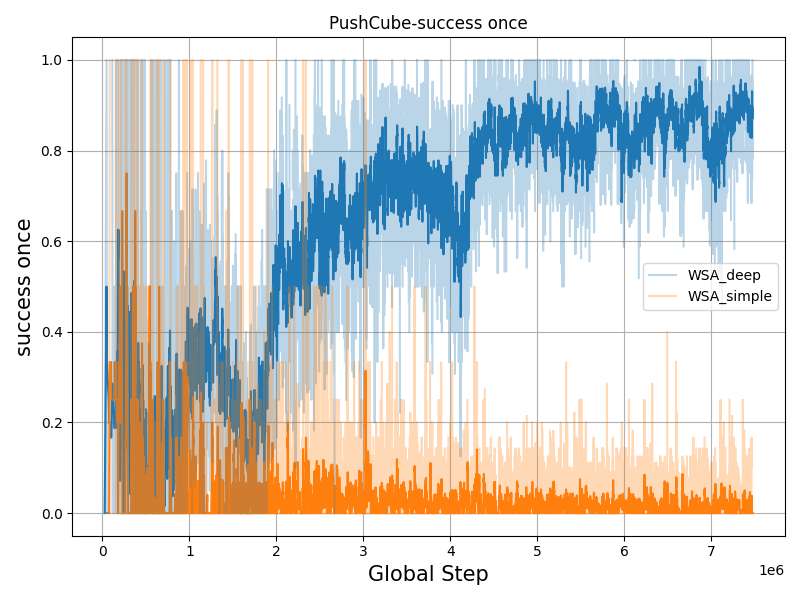}
        \caption{WSA agent success once}
        \label{fig:pushcube success}
    \end{subfigure}
    \caption{Training performance of WSA agent on \texttt{Push-Cube-v1} environment, reporting episode reward and success during training. Comparing performance of WSA using a single layer (\textit{simple}) or three layer (\textit{deep}) \texttt{Shared Weight Network}.}
    \label{fig:pushcube training}
\end{figure}

\begin{figure}[ht]
    \centering
    \begin{subfigure}[b]{0.40\textwidth}
        \includegraphics[width=0.9\textwidth]{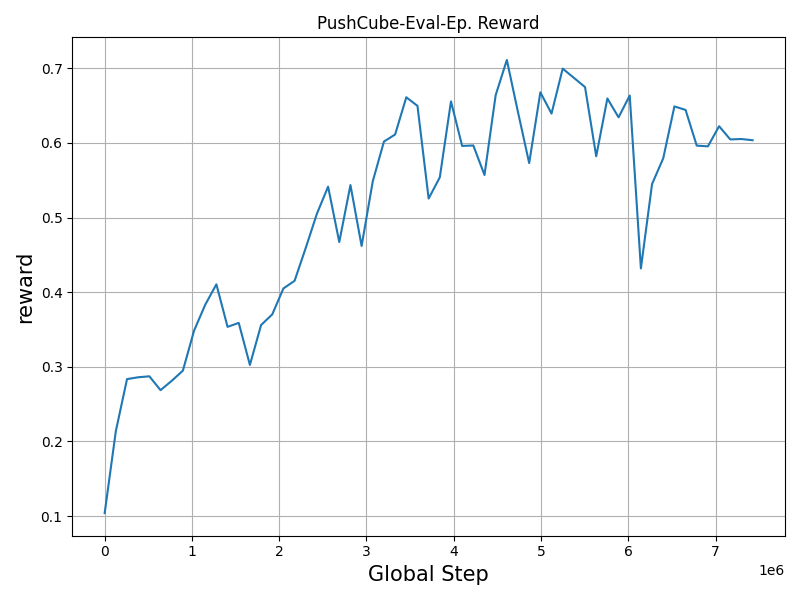}
        \caption{deep WSA mean reward on evaluation}
        \label{fig:pushcube eval rew}
    \end{subfigure}
    \begin{subfigure}[b]{0.40\textwidth}
            \includegraphics[width=0.9\textwidth]{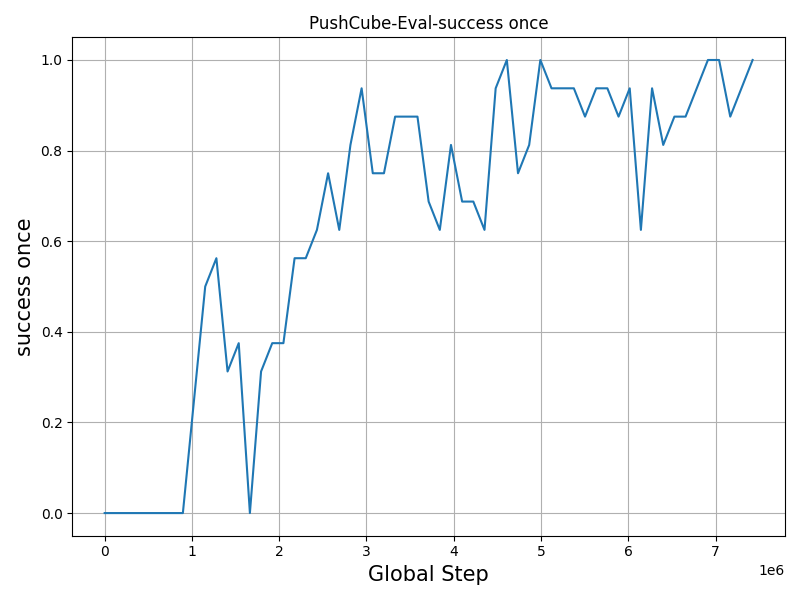}
        \caption{deep WSA success once on evaluation}
        \label{fig:pushcube eval success}
    \end{subfigure}
    \caption{Evaluation of deep WSA agent on \texttt{Push-Cube-V1}. Evaluation is performed over 16 episodes.}
    \label{fig:pushcube eval}
\end{figure}

In this experiment we consider the \texttt{Push-Cube-v1} environment with an episode length of 50, where the robot arm has to move a block to a target location. We use the RGB representation to define the observation space and use PPO as RL algorithm. To process the WSA's visual representation, we use the following pre-trained models: (\textit{i}) \texttt{SwinTransformer} \citep{liu2021swin}; (\textit{ii}) a fine-tuned \texttt{resnet18} from R3M \citep{nair2022r3muniversalvisualrepresentation} tailored for robotic tasks; (\textit{iii}) a \texttt{CLIP} model \citep{radford2021learningtransferablevisualmodels} with a ViT backbone \citep{dosovitskiy2021imageworth16x16words} for the state representation. Among these pre-trained models, only \texttt{resnet18} was already fine-tuned for robotics tasks, while the others remained general-purpose pre-trained vision encoders. In contrast to the discrete action spaces of Atari environments, this domain requires continuous control, making the task substantially more challenging. In fact, WSA default architecture using single-layer MLP for the \texttt{Shared Weight Network} collapses and fails to learn a meaningful policy. To better handle the complexity of the problem, we deepened the architecture of the \texttt{Shared Weight Network} to \textit{three} hidden layers, keeping its size relatively small. Although this setting is substantially more complex, a minimal architectural adjustment allows WSA to achieve strong performance. Figure \ref{fig:pushcube training} shows the training curves for both reward and success, comparing the basic and deep variants of WSA. While we fixed the number of training steps to 7.5M, the performance increases steadily as training progresses, suggesting that with longer training time, WSA can converge to optimal performance as shown in Figure \ref{fig:pushcube eval}, where deep WSA during evaluation on 16 episodes reaches a mean reward near \textbf{0.6} per step and success rate close to \textbf{90\%}. This experiment demonstrate that our architecture generalizes effectively to diverse environments, including those more realistic and complex than Atari games, achieving strong performance with minimal tuning.

To widen the comparison introduced in Section \ref{sec:related_work} between WSA and recent approaches, we extend our evaluation to include the additional robotic scenario from \texttt{ManiSkill}. In this extended analysis, we incorporate two representative methods: InstructRL \citep{liu2022instruction} and OpenVLA \citep{kim2024openvla}.
InstructRL is composed of two main modules: (\textit{i}) a pre-trained multi-modal masked autoencoder (M3AE) \citep{geng2022multimodal}, and (\textit{ii}) a transformer-based policy. We use the publicly available pre-trained encoder weights and modify the code to integrate it with CleanRL \citep{cleanrl}. This allows us to train the RL agent online, as described in this paper's experimental setup.
By contrast, OpenVLA is a 7B-parameter open-source vision-language-action model (VLA) trained on robot trajectories from the Open X-Embodiment dataset \citep{o2024open}. Its architecture combines (\textit{i}) a vision-encoder module that fuses DinoV2 \citep{oquab2023dinov2} and SigLIP \citep{zhai2023sigmoid} representations via a MLP, with (\textit{ii}) a Llama 2 7B-parameter large language model backbone \citep{touvron2023llama}. Following the authors’ instructions, we downloaded the pre-trained OpenVLA checkpoint from HuggingFace and fine-tuned it specifically on the ManiSkill dataset, available on the Open X-Embodiment dataset, following the provided scripts.

The results, reported in Table \ref{tab:method_comparison}, highlight two complementary aspects: (\textit{i}) the ability of each method to solve the \texttt{Push-Cube} task, and (\textit{ii}) the computational resources required during training or fine-tuning. We remind that InstructRL is evaluated in an online training setting, mirroring the setup used in this work, while OpenVLA is trained offline following the authors’ publicly available fine-tuning procedure. Despite using the same number of interactions (7.5 million steps), InstructRL is slower than WSA, primarily due to its architecture, which involves a masked multi-model autoencoder followed by a transformer policy. Additionally, after 7.5M training steps, InstructRL fails to converge and learn meaningful behavior, resulting in a low average reward per step and no success during evaluation. Instead, by leveraging representations from general pre-trained models, WSA learns an effective policy that achieves a 90\% success rate during evaluation under the same interaction budget, while reducing training time by a factor of \textbf{2.8}. These results suggest that InstructRL's proposed architecture either requires substantially more training steps to converge in an online setting or is better suited to the offline paradigm used in its original work. On the other hand, OpenVLA’s fine-tuning process, executed with the authors’ script using LoRA \citep{hu2022lora} on a single A100 GPU, demands even greater computational investment. Although OpenVLA’s training metrics showed improved action prediction on ManiSkill task data, its evaluation performance remained poor. Throughout the evaluation, we observed that the agent repeatedly stalled near the cube, yielding a 0\% success rate regardless of prompt style. Moreover, OpenVLA’s 50k fine-tuning steps took approximately 92 hours to complete, which is close to \textbf{6.5x} the time required by WSA. Its large 7 billion–parameter VLA backbone leads to heavy compute cost per iteration. WSA’s compact architecture not only allows the agent to learn an effective policy within a modest time budget but also imposes far fewer specialized dependencies and hyperparameter tweaks. Under the same experimental configuration, InstructRL failed to converge, whereas WSA achieved a 90\% success rate. Consequently, our proposed architecture emerges as a more flexible and readily adaptable solution. In this setting, where the \textit{only} change from the Atari baseline was the addition of two hidden layers to the \texttt{Shared Weight Network}, WSA strikes a superior balance between training efficiency, computational cost, and final performance.

\begin{table}[ht]
    \centering
    \begin{tabular}{l|cccc}
        \textbf{Method} & \textbf{Reward/Step} & \textbf{Success Rate} & \textbf{Training Steps} & \textbf{Training Time} \\
        \addlinespace
        \hline
        \addlinespace
        WSA (\textit{Ours}) & 0.60 & 0.91 & 7.5M & 14 h \\
        InstructRL \citep{liu2022instruction} & 0.14 & 0.0 & 7.5M & 40 h \\
        OpenVLA \citep{kim2024openvla} & 0.10 & 0.0 & 50K & 92 h \\
    \end{tabular}
    \caption{Performance comparison during evaluation on \texttt{Push-Cube-V1}. All models are run on a single A100 GPU.}
    \label{tab:method_comparison}
\end{table}

\pagebreak

\subsection{Breakout Additional Experiments}\label{sec:breakout-add}
\begin{table}[ht]
\begin{minipage}[b]{0.5\linewidth}
A first attempt to overcome the problem was to increase the number of parameters of the model, adding more expressive power to the network learning the policy. We increased the size of the \texttt{Fully-Connected Network} to three linear layers, using ReLU as activation function. Figure \ref{fig:break_policy_new} and Table \ref{tab:results_breakout_extra} - referenced as \texttt{(P)} - report the results for this experiment. With respect to the default scenario, there is an improvement in performance, but WSA is still far from the E2E model (\texttt{156 vs 404)}.\\
The additional analysis, reported in Section \ref{sec:breakout_study}, was also extended to the other two combination modules studied for the game Breakout (Section \ref{sec:init_exp} and Figure \ref{fig:breakouttrain}). Figures \ref{fig:breakout_cnn_study}, \ref{fig:breakout_fix_study} showcase the training trend while Table \ref{tab:results_breakout_extra} evaluation scores (we recall \texttt{(P)} indicates that the  \texttt{(Fully-Connected Network)} was increased to more layers, while \texttt{(M)} show the result while re-training the models using both random and expert data). Differently from WSA, these modules do not scale as good their final performance, they are still far from PPO final score. Nevertheless, \texttt{CNN} seems to gain advantage from an increase amount of hyperparameter, achieving the highest score in the \texttt{Policy} (P) version, while \texttt{FIX} shows a similar trend of improvement that was observed with WSA.
\end{minipage}\hfill
\begin{minipage}[b]{0.45\linewidth}
\centering
\includegraphics[width=0.9\textwidth]{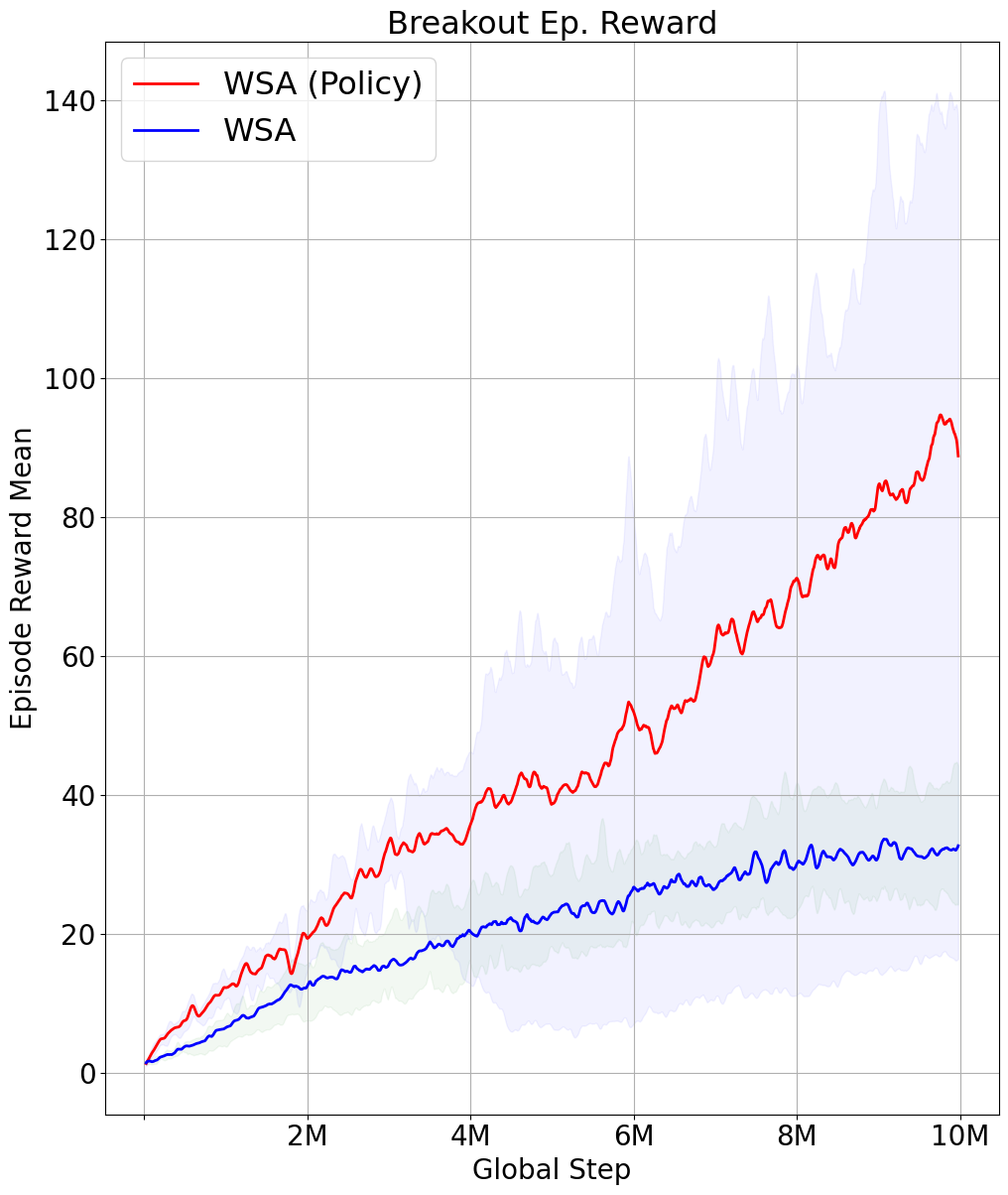}
\captionof{figure}{Increasing MLP size}
\label{fig:break_policy_new}
\end{minipage}
\end{table}

\begin{figure}[ht]
    \centering
    \hfill
    \begin{subfigure}[b]{0.45\textwidth}
        \centering
        \includegraphics[width=0.6\textwidth]{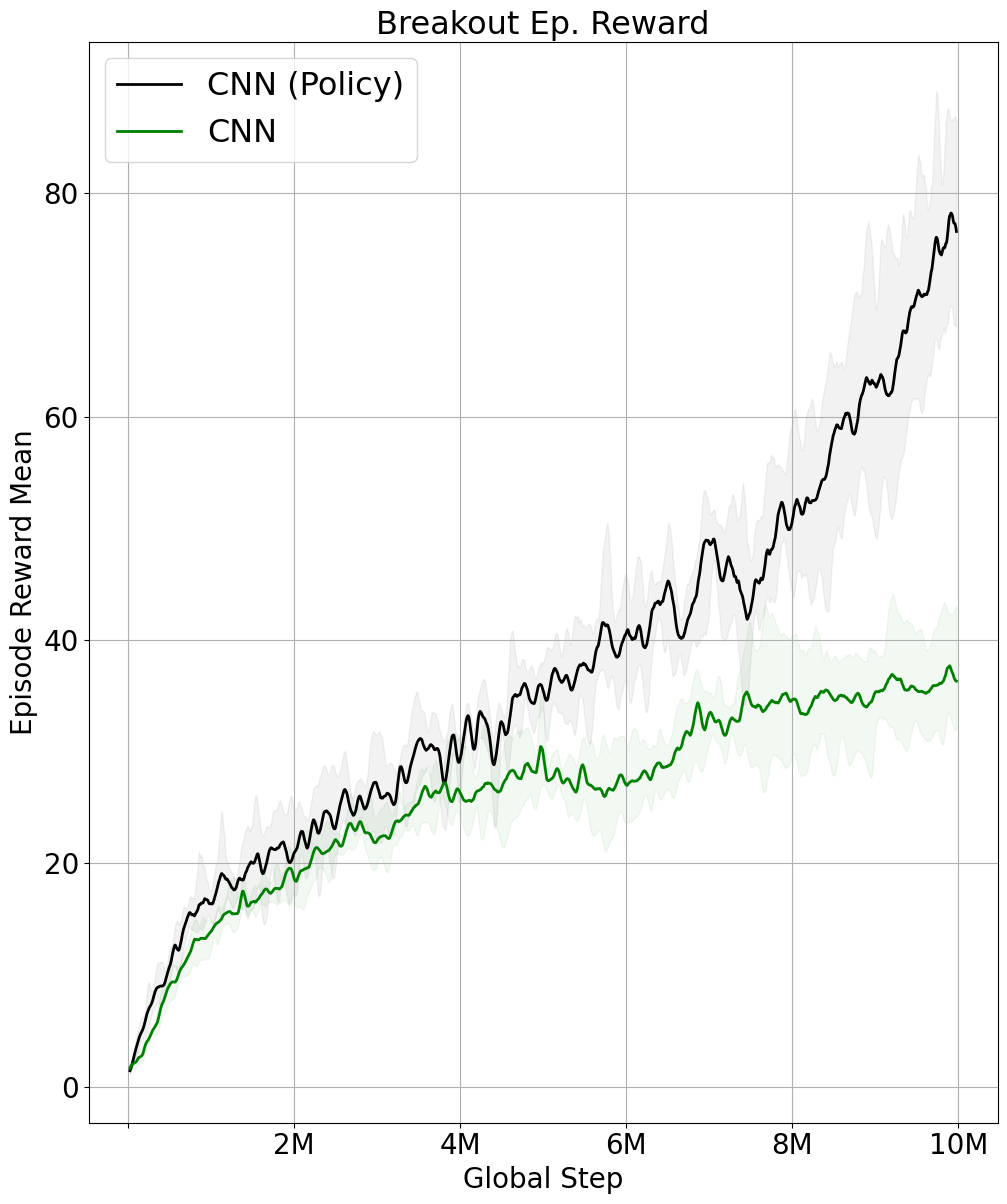}
        \caption{Increasing MLP size}
        \label{fig:breakout_cnn_policy_policy}
    \end{subfigure}
    \hfill
    \begin{subfigure}[b]{0.45\textwidth}
        \centering
        \includegraphics[width=0.6\textwidth]{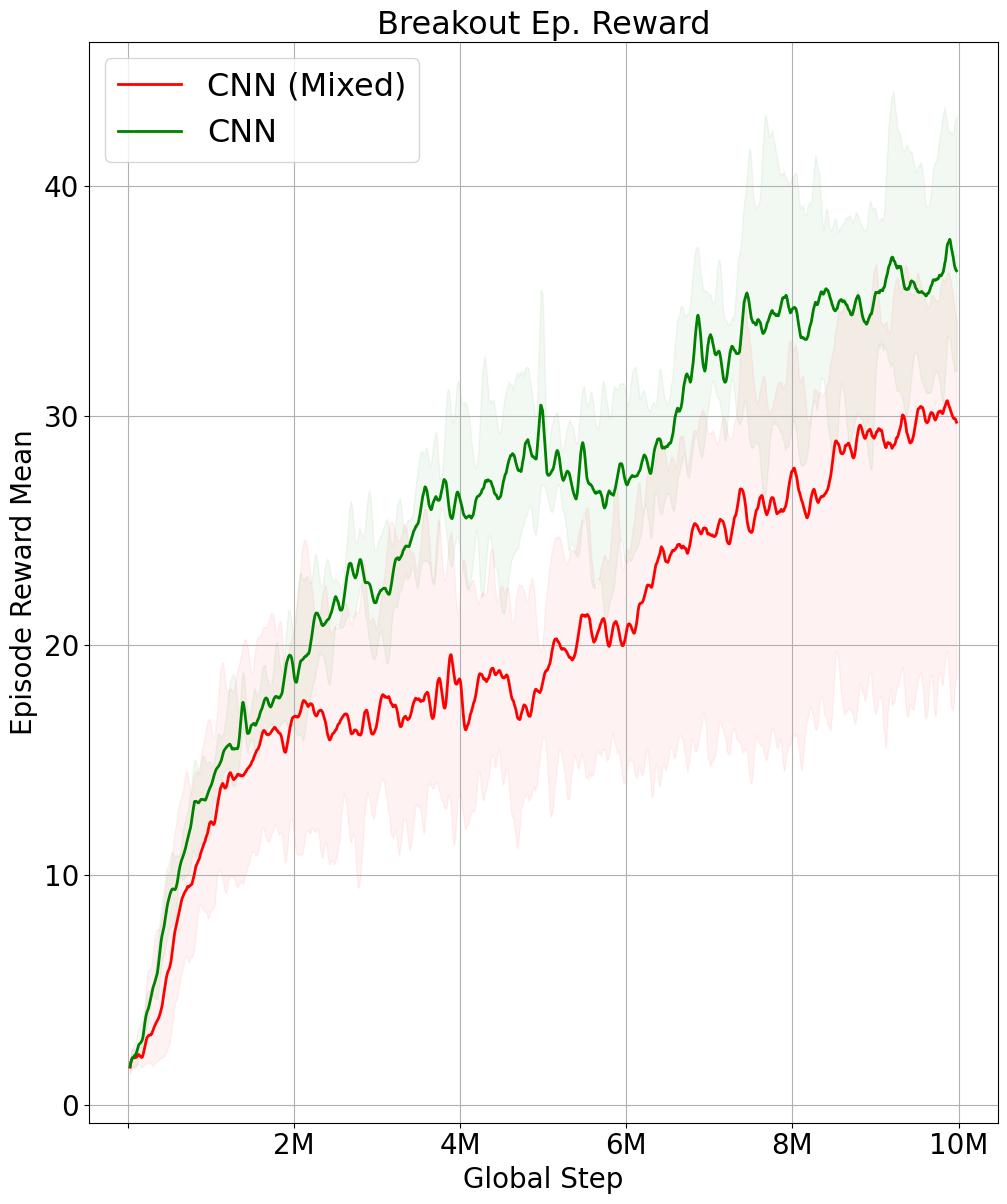}
        \caption{Using Mixed Data}
        \label{fig:breakout_cnn_mixed}
    \end{subfigure}
    \caption{Performance comparison of \texttt{CNN} with PPO across various strategies in \texttt{Breakout}. Each subfigure displays the average score with the standard deviation shaded.}
\label{fig:breakout_cnn_study}
\end{figure}
\begin{figure}[ht]
    \centering
    \begin{subfigure}[b]{0.45\textwidth}
        \centering
        \includegraphics[width=0.6\textwidth]{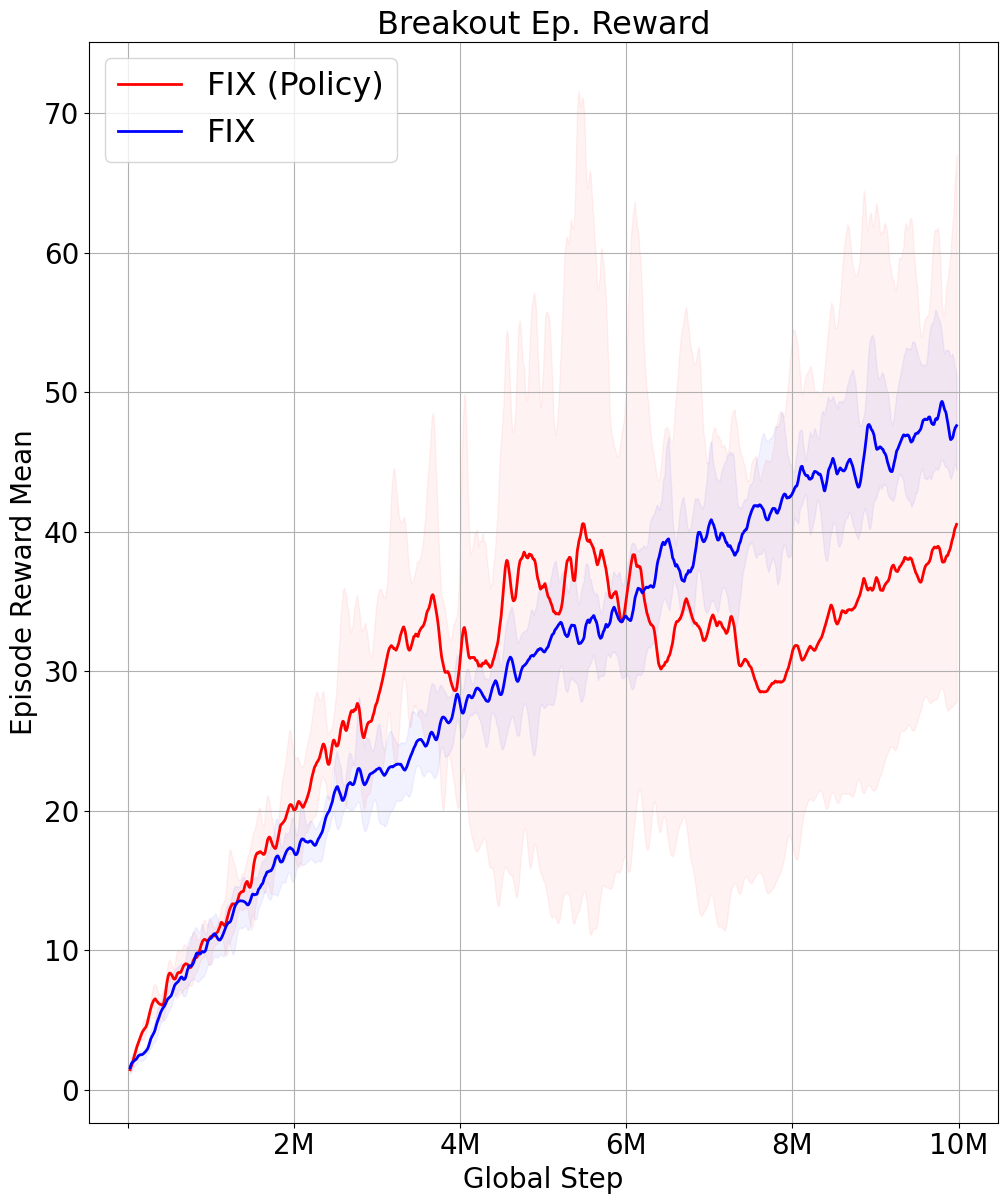}
        \caption{Increasing MLP size}
        \label{fig:breakout_fix_policy_policy}
    \end{subfigure}
    \hfill
    \begin{subfigure}[b]{0.45\textwidth}
        \centering
        \includegraphics[width=0.6\textwidth]{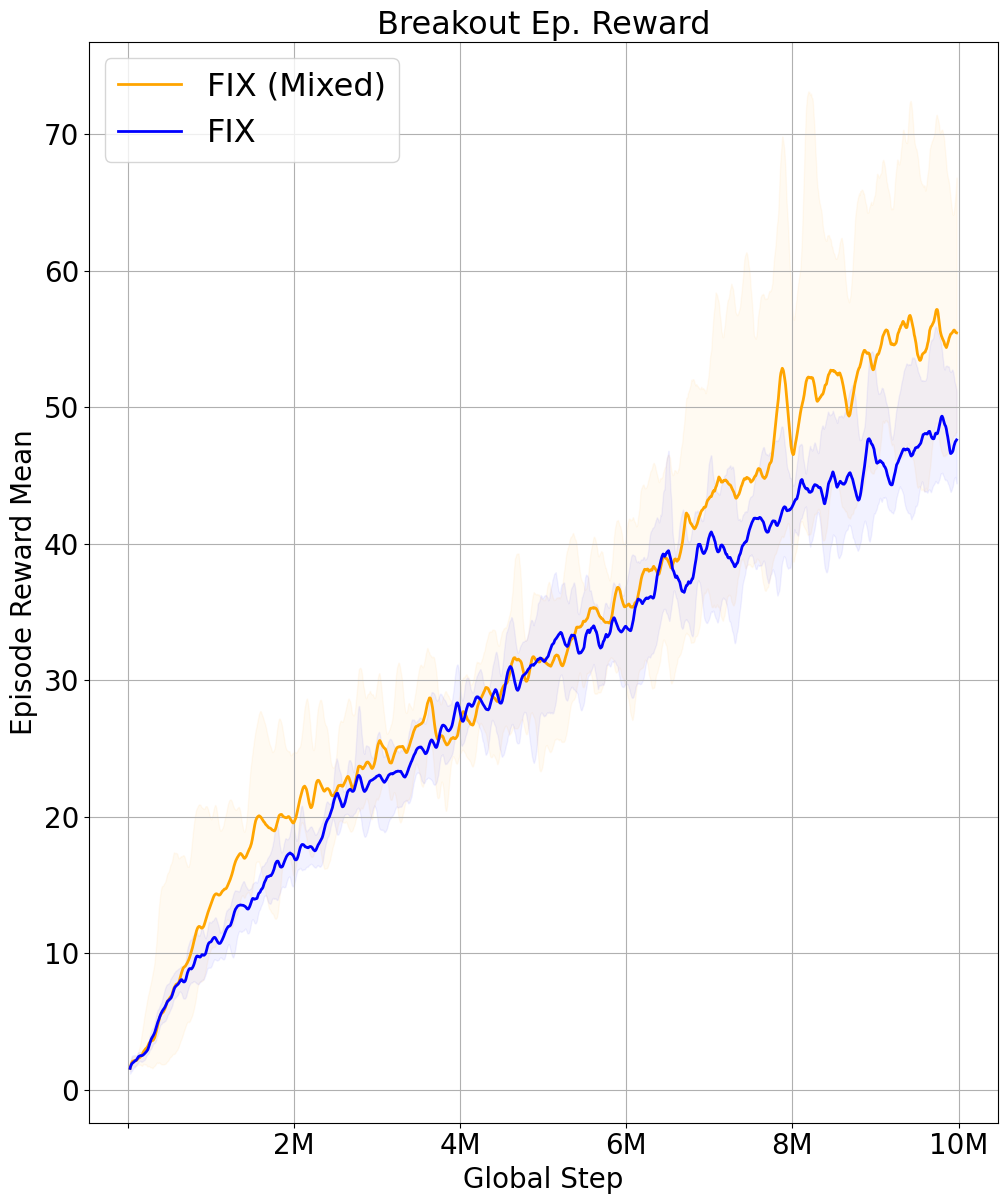}
        \caption{Using Mixed Data}
        \label{fig:breakout_fix_mixed}
    \end{subfigure}
    \caption{Performance comparison of \texttt{FIX} with PPO across various strategies in \texttt{Breakout}. Each subfigure displays the average score with the standard deviation shaded.}
    \label{fig:breakout_fix_study}
\end{figure}

\begin{table}[!ht]
    \begin{center}
        \begin{tabular}{lll}
            \multicolumn{1}{l}{}  &\multicolumn{1}{l}{\bf Agent} &\multicolumn{1}{l}{\bf Reward} 
            \\ \hline \\           
            \multirow{9}{*}{\texttt{Breakout}}
                                  & CNN & 65.98 $\pm$ 1.62 \\
                                  & CNN \texttt{(P)} & 118.71 $\pm$ 4.30 \\
                                  & CNN \texttt{(M)} & 62.21 $\pm$ 1.98 \\
                                  & FIX & 87.17 $\pm$ 6.87 \\
                                  & FIX \texttt{(P)} & 106.46 $\pm$ 10.84 \\
                                  & FIX \texttt{(M)} & 199.08 $\pm$ 7.46\\
                                  & WSA \texttt{(P)} & 156.17 $\pm$ 3.59 \\
                                  & E2E & 404.46 $\pm$ 13.49 \\
        \end{tabular}
    \end{center}
    \caption{Performance during \texttt{evaluation}. This completes the results reported in Table \ref{tab:results}.}
    \label{tab:results_breakout_extra}
\end{table}

\pagebreak

\section{HackAtari Full Evaluation}\label{sec:app-hackatari}
\begin{figure}[ht]
    \centering
    \begin{subfigure}[b]{0.32\textwidth}
        \includegraphics[width=\textwidth]{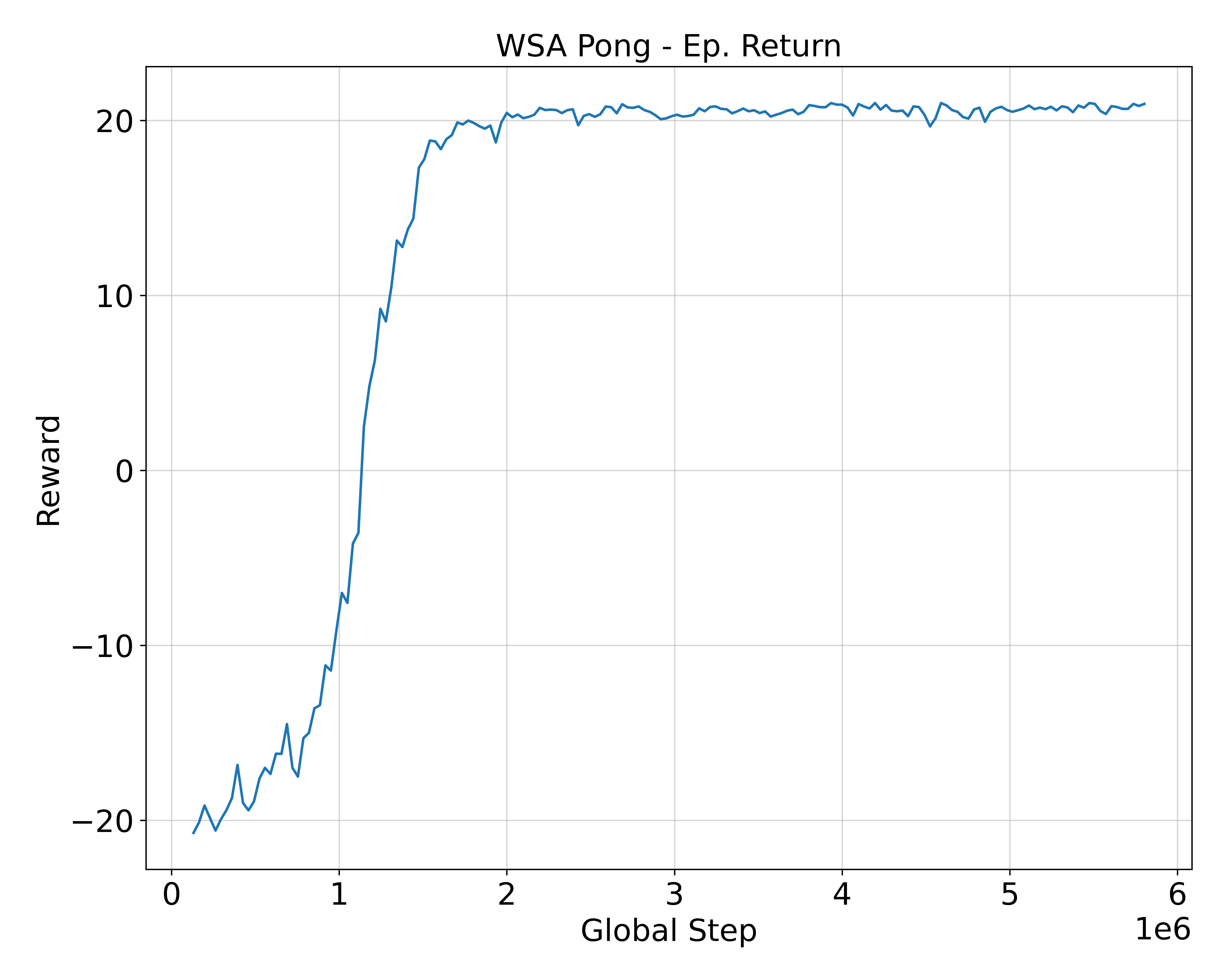}
        \caption{WSA Pong}
        \label{fig:pong_hack_train}
    \end{subfigure}
    \hfill
    \begin{subfigure}[b]{0.32\textwidth}
        \centering
        \includegraphics[width=\textwidth]{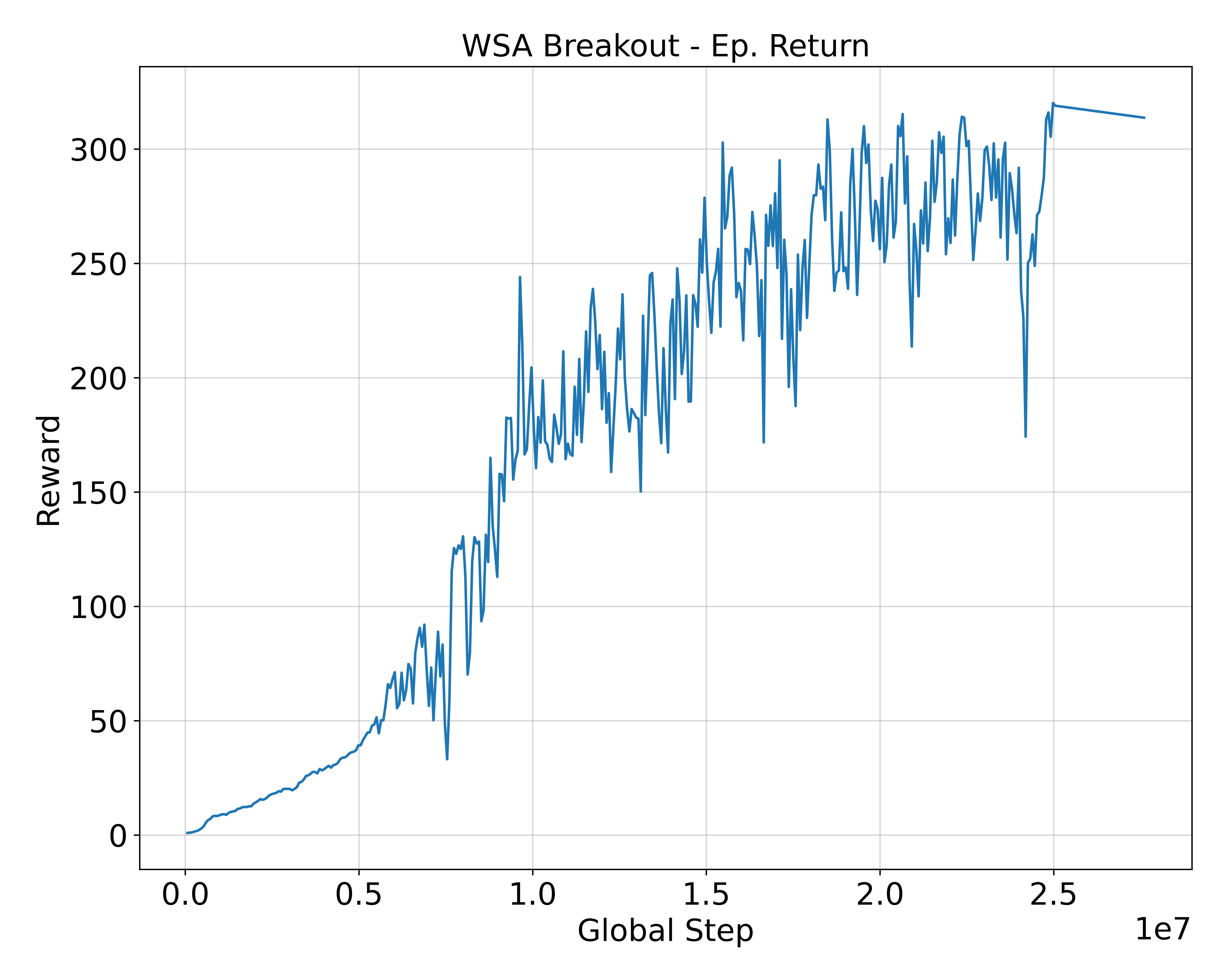}
        \caption{E2E Breakout}
        \label{fig:breakout_hack_train_ppo}
    \end{subfigure}
    \hfill
    \begin{subfigure}[b]{0.32\textwidth}
        \centering
        \includegraphics[width=\textwidth]{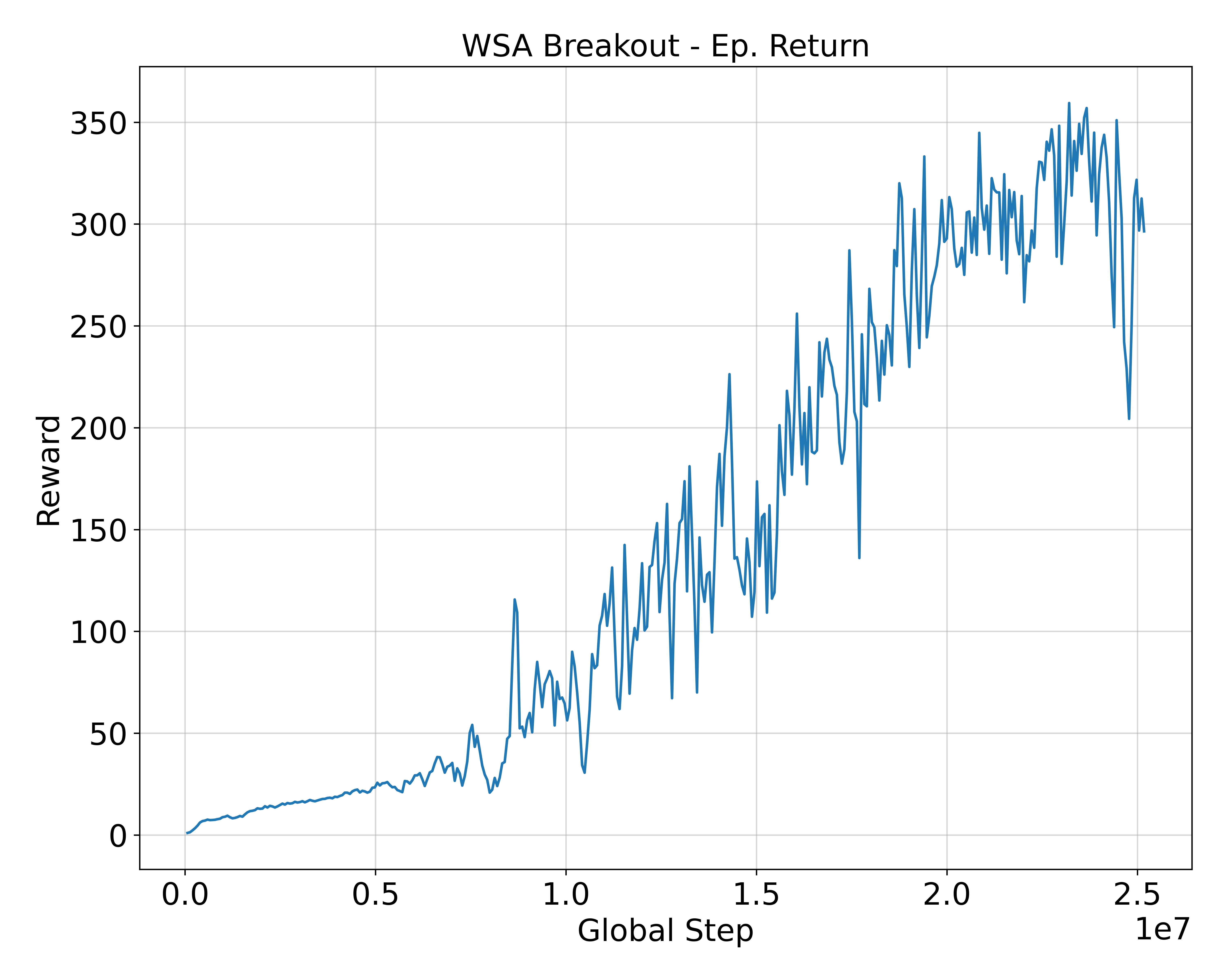}
        \caption{WSA Breakout}
        \label{fig:breakout_hack_train_wsa}
    \end{subfigure}
    \caption{Training curves of episodes reward for the agents used in the HackAtari evaluations on Pong and Breakout.}
    \label{fig:hack_training}
\end{figure}

In this section, we report the training curves of the three models we trained to be evaluated in the HackAtari scenarios. Figures \ref{fig:pong_hack_train}-\ref{fig:breakout_hack_train_wsa} outline the training curves for the WSA agents on Pong, reaching over 20 as the final score, and Breakout getting close to 300 points. Since in \citet{hackatari} no information on the performance of PPO on Breakout variations is provided, we also train an end-to-end (E2E) PPO agent matching the training steps of WSA (Figure \ref{fig:breakout_hack_train_ppo}). Table \ref{tab:breakout_hack_full} details the performance of PPO and WSA across all 25 possible variations of Breakout. We recall that CP and CB stand for \textit{color\_player} and \textit{color\_blocks}, respectively. Both can range between five different values $X \in [0, 4]$ matching the colors black, white, red, blue, and green.

\begin{table}[ht]
    \centering
    \begin{tabular}{c|c|c|c}
        Game & Random &E2E & WSA\\
        \addlinespace
        \hline 
        \addlinespace
         \begin{tabular}{cc} 
            \textbf{Training} \\
            \textbf{Testing}
         \end{tabular} &
         \begin{tabular}{cc} 
            - \\
            variation
        \end{tabular} &
        \begin{tabular}{cc} 
            original \\
            variation
        \end{tabular} &
        \begin{tabular}{cc}
            original \\
            variation
        \end{tabular} \\
        \addlinespace
        \hline
        \addlinespace
        Breakout (CP 0) (CB 0) & $0.98 \pm 1.05$ & $12.58 \pm 7.91$ & $44.92 \pm 18.27$ \\
        Breakout (CP 0) (CB 1) & $1.40 \pm 1.22$ & $12.28 \pm 5.67$ & $46.62 \pm 25.13$ \\
        Breakout (CP 0) (CB 2) & $1.36 \pm 1.51$ & $11.14 \pm 7.11$ & $43.48 \pm 14.19$ \\
        Breakout (CP 0) (CB 3) & $0.92 \pm 1.04$ & $11.98 \pm 6.00$ & $47.10 \pm 15.18$ \\
        Breakout (CP 0) (CB 4) & $1.32 \pm 1.27$ & $12.76 \pm 10.01$ & $46.08 \pm 19.63$ \\
        \addlinespace
        Breakout (CP 1) (CB 0) & $0.96 \pm 0.92$ & $13.04 \pm 7.46$ & $50.06 \pm 18.65$ \\
        Breakout (CP 1) (CB 1) & $1.10 \pm 1.17$ & $14.66 \pm 12.33$ & $43.28 \pm 16.37$ \\
        Breakout (CP 1) (CB 2) & $1.18 \pm 1.13$ & $13.5 \pm 9.99$ & $43.96 \pm 17.17$ \\
        Breakout (CP 1) (CB 3) & $1.32 \pm 1.42$ & $13.48 \pm 7.23$ & $44.56 \pm 17.16$ \\
        Breakout (CP 1) (CB 4) & $1.36 \pm 1.20$ & $13.06 \pm 8.93$ & $44.90 \pm 16.68$ \\
        \addlinespace
        Breakout (CP 2) (CB 0) & $1.54 \pm 1.34$ & $12.92 \pm 7.68$ & $45.50 \pm 15.40$ \\
        Breakout (CP 2) (CB 1) & $1.12 \pm 1.11$ & $13.46 \pm 8.84$ & $44.56 \pm 19.90$ \\
        Breakout (CP 2) (CB 2) & $1.40 \pm 1.23$ & $13.40 \pm 9.18$ & $43.60 \pm 16.50$ \\
        Breakout (CP 2) (CB 3) & $1.26 \pm 1.35$ & $12.92 \pm 10.40$ & $43.46 \pm 16.69$ \\
        Breakout (CP 2) (CB 4) & $1.26 \pm 1.18$ & $12.68 \pm 9.02$ & $43.10 \pm 14.02$ \\
        \addlinespace
        Breakout (CP 3) (CB 0) & $0.98 \pm 1.12$ & $13.42 \pm 8.76$ & $39.06 \pm 16.54$ \\
        Breakout (CP 3) (CB 1) & $0.98 \pm 1.12$ & $12.8 \pm 7.91$ & $45.88 \pm 15.49$ \\
        Breakout (CP 3) (CB 2) & $0.96 \pm 0.94$ & $12.72 \pm 8.24$ & $45.84 \pm 14.17$ \\
        Breakout (CP 3) (CB 3) & $0.96 \pm 1.06$ & $13.12 \pm 8.83$ & $44.38 \pm 17.33$ \\
        Breakout (CP 3) (CB 4) & $1.14 \pm 1.10$ & $13.44 \pm 10.23$ & $42.30 \pm 17.72$ \\
        \addlinespace
        Breakout (CP 4) (CB 0) & $1.46 \pm 1.28$ & $12.10 \pm 7.02$ & $42.86 \pm 16.69$ \\
        Breakout (CP 4) (CB 1) & $1.16 \pm 1.22$ & $13.38 \pm 11.42$ & $43.42 \pm 16.81$ \\
        Breakout (CP 4) (CB 2) & $1.32 \pm 1.27$ & $11.72 \pm 5.50$ & $47.80 \pm 21.65$ \\
        Breakout (CP 4) (CB 3) & $0.90 \pm 1.08$ & $12.66 \pm 7.85$ & $46.52 \pm 18.12$ \\
        Breakout (CP 4) (CB 4) & $0.96 \pm 0.96$ & $12.2 \pm 6.77$ & $42.94 \pm 15.29$ \\
    \end{tabular}
    \caption{Evaluations over 30 episodes, reporting mean and std, analyzing the performance over all possible variations of Breakout.}
    \label{tab:breakout_hack_full}
\end{table}

\section{CARBS Hyperparameters}\label{sec:app-carbs}
The hyperparameter search for the model training is performed using CARBS \citep{carbs}, which systematically explores different combinations of hyperparameters to optimize performance. Table \ref{tab:hyperparams} reports the ranges for the key hyperparameters used in the search.

\begin{table}[h]
    \centering
    \begin{tabular}{l|c}
        \textbf{Hyperparameter} & \textbf{Range} \\
        \addlinespace
        \hline
        \addlinespace
        \texttt{Batch Size} & 16,384 -- 65,536 \\
        \texttt{BPTT Horizon} & $\{1, 2, 4, 8, 16\}$ \\
        \texttt{Clip Coefficient} & 0.0 -- 1.0 \\
        \texttt{Embedding Size} & 128 -- 1,024 \\
        \texttt{Entropy Coefficient} & $1e-5$ -- $1e-1$ \\
        \texttt{GAE Lambda} & 0.0 -- 1.0 \\
        \texttt{Gamma} & 0.0 -- 1.0 \\
        \texttt{Learning Rate} & $1e-5$ -- $1e-1$ \\
        \texttt{Max Gradient Norm} & 0.0 -- 10.0 \\
        \texttt{Minibatch Size} & 512 -- 2,048 \\
        \texttt{Total Timesteps} & 5M -- 25M \\
        \texttt{Update Epochs} & 1 -- 4 \\
        \texttt{Value Function Clip Coefficient} & 0.0 -- 1.0 \\
        \texttt{Value Function Coefficient} & 0.0 -- 1.0 \\
    \end{tabular}
    \caption{Hyperparameter ranges used for the search.}
    \label{tab:hyperparams}
\end{table}

\section{Combination Modules Analysis} \label{sec:app-com-mod}
In this section, we provide additional insights for the results of the initial analysis. The experimental setup, outlined in Sections \ref{sec:init_exp} and Appendix \ref{sec:app-exp-setup}, structures a robust framework for evaluating the various combination modules.

\begin{figure}[h!]
    \centering
    \begin{subfigure}[b]{0.32\textwidth}
        \includegraphics[width=\textwidth]{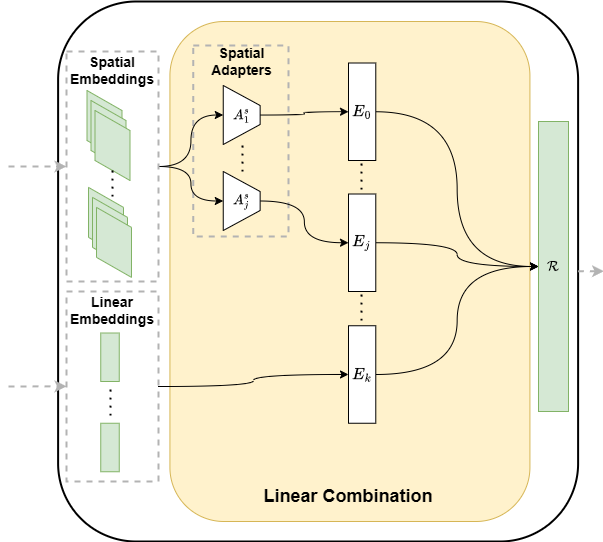}
        \caption{\texttt{LIN}}
    \end{subfigure}
    \hfill
    \begin{subfigure}[b]{0.32\textwidth}
        \centering
        \includegraphics[width=\textwidth]{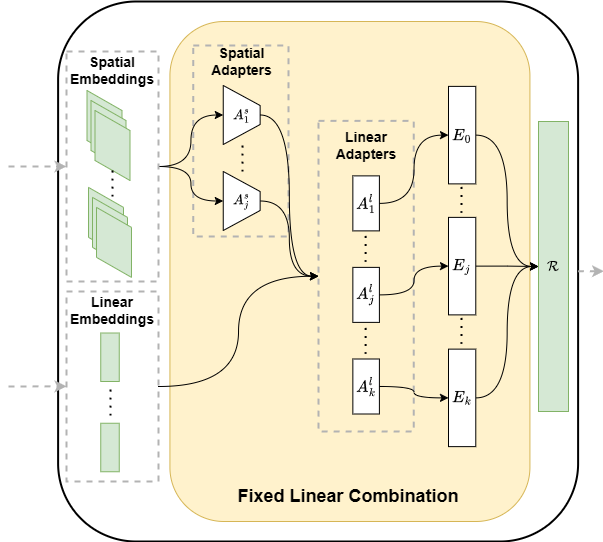}
        \caption{\texttt{FIX}}
    \end{subfigure}
    \hfill
    \begin{subfigure}[b]{0.32\textwidth}
        \centering
        \includegraphics[width=\textwidth]{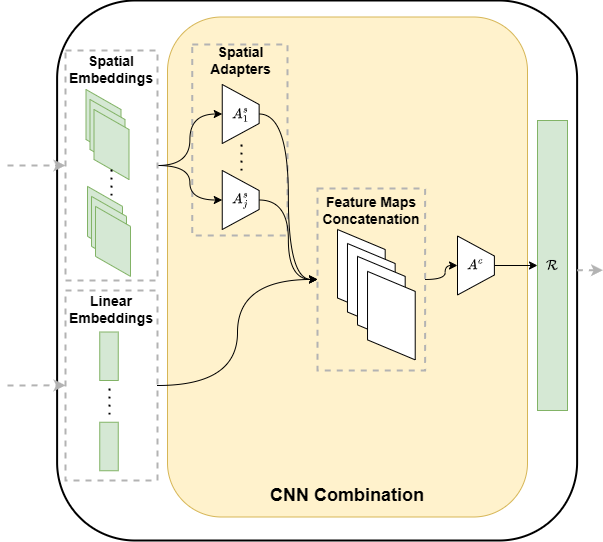}
        \caption{\texttt{CNN}}
    \end{subfigure}
    \caption{Graphical representation of different embedding combination modes that have been evaluated. The blocks representing the module can be placed in Figure \ref{fig:main_architecture} replacing WSA.}
    \label{fig:com1}
\end{figure}

\begin{figure}[h!]
    \centering
    \begin{subfigure}[b]{0.32\textwidth}
        \includegraphics[width=\textwidth]{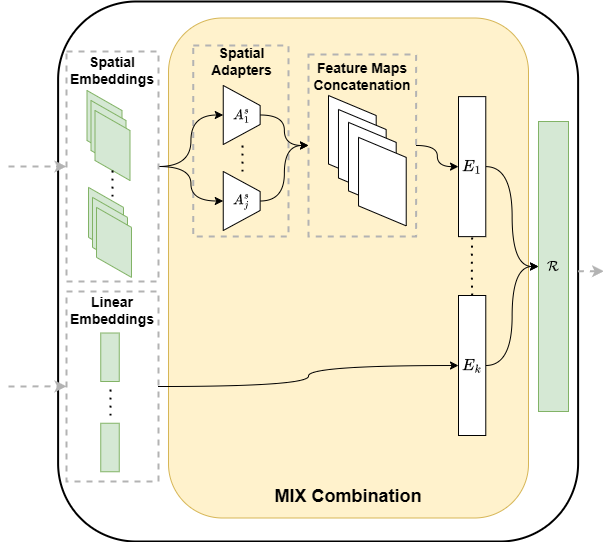}
        \caption{\texttt{MIX}}
    \end{subfigure}
    \hfill
    \begin{subfigure}[b]{0.32\textwidth}
        \centering
        \includegraphics[width=\textwidth]{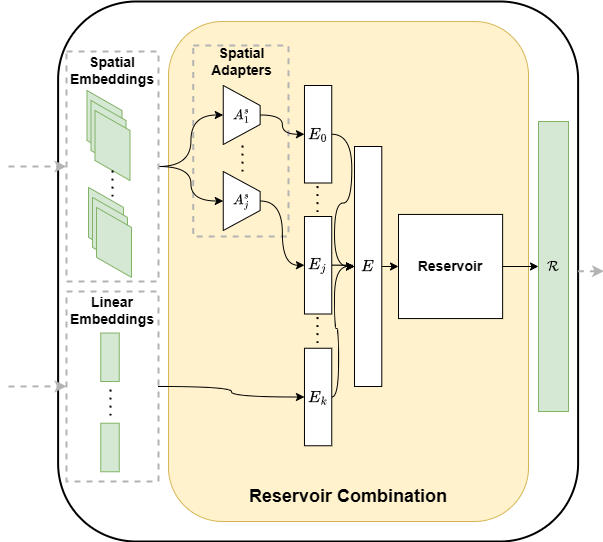}
        \caption{\texttt{RES}}
    \end{subfigure}
    \hfill
    \begin{subfigure}[b]{0.32\textwidth}
        \centering
        \includegraphics[width=\textwidth]{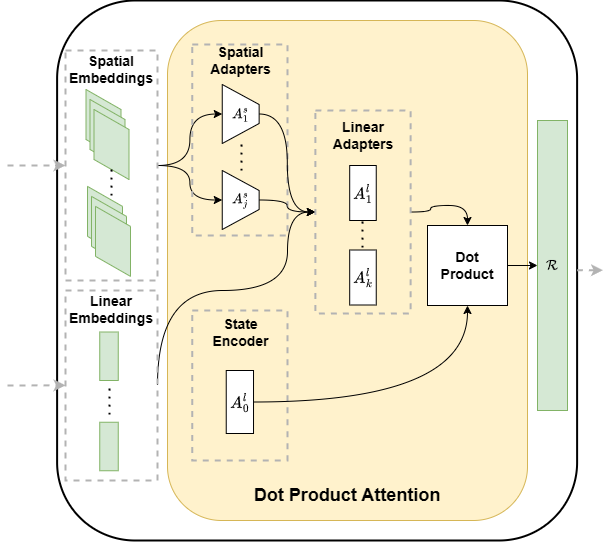}
        \caption{\texttt{DPA}}
    \end{subfigure}
    \caption{Graphical representation of different embedding combination modes that have been evaluated. The blocks representing the module can be placed in Figure \ref{fig:main_architecture} replacing WSA.}
    \label{fig:com2}
\end{figure}

\begin{figure}[ht]
    \centering
    \begin{subfigure}[b]{0.45\textwidth}
        \centering
        \includegraphics[width=\textwidth]{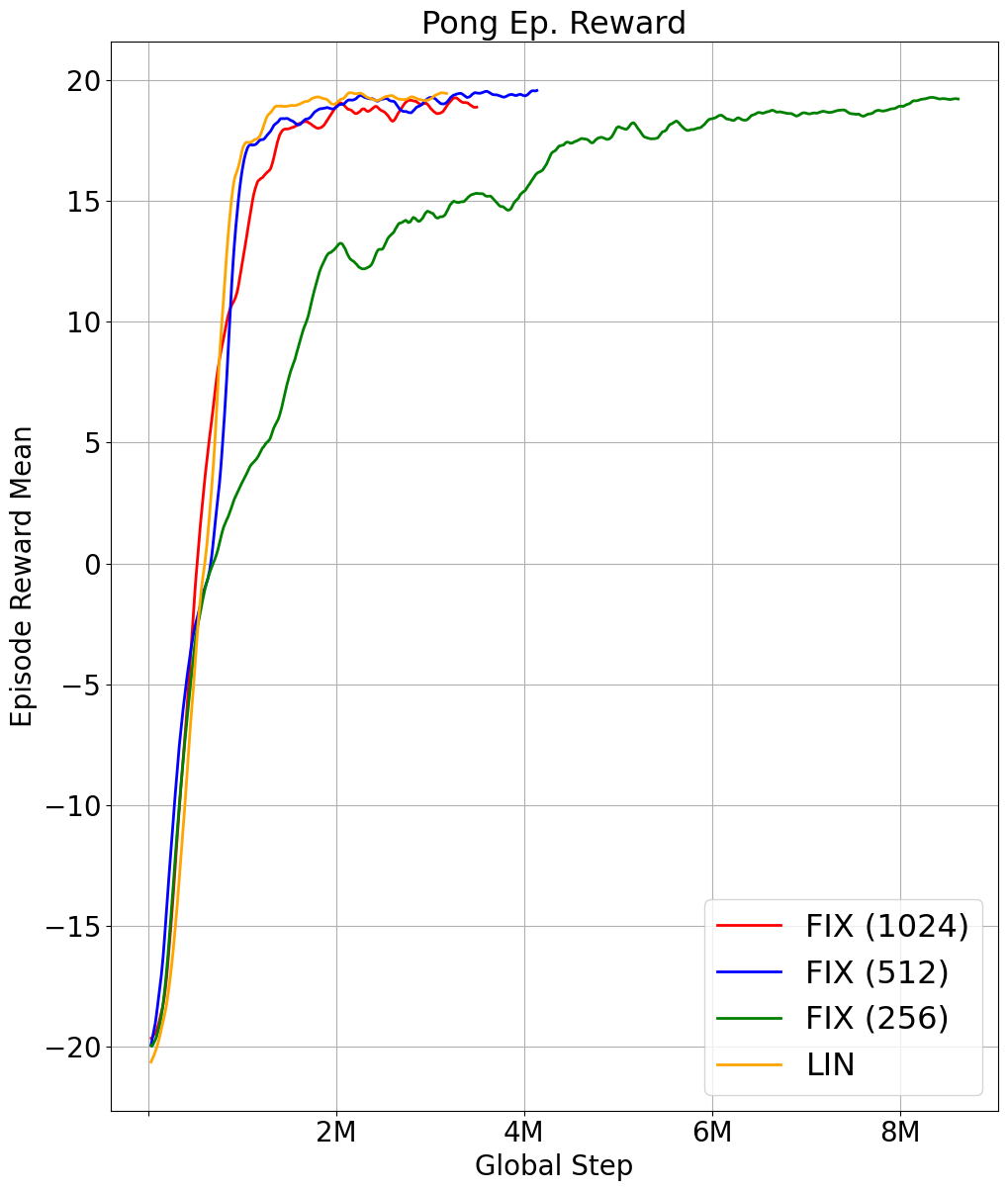}
        \caption{\texttt{Linear} and \texttt{Fixed Linear}}
        \label{fig:pong_lin_fix}
    \end{subfigure}
    \hfill
    \begin{subfigure}[b]{0.45\textwidth}
        \centering
        \includegraphics[width=\textwidth]{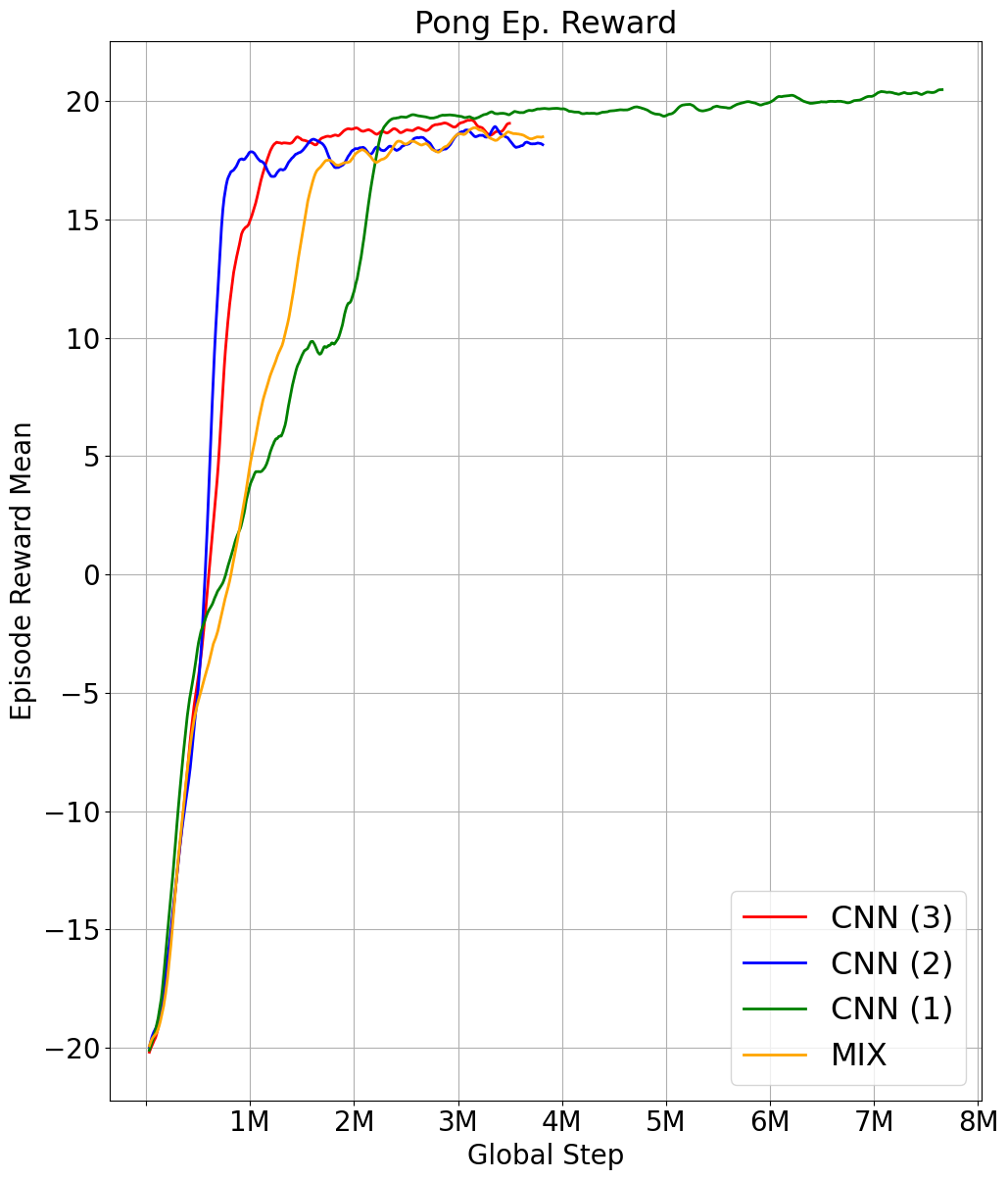}
        \caption{\texttt{Convolutional} and \texttt{Mixed}}
        \label{fig:pong_cnn_mixed}
    \end{subfigure}
    \hfill
    \begin{subfigure}[b]{0.45\textwidth}
        \centering
        \includegraphics[width=\textwidth]{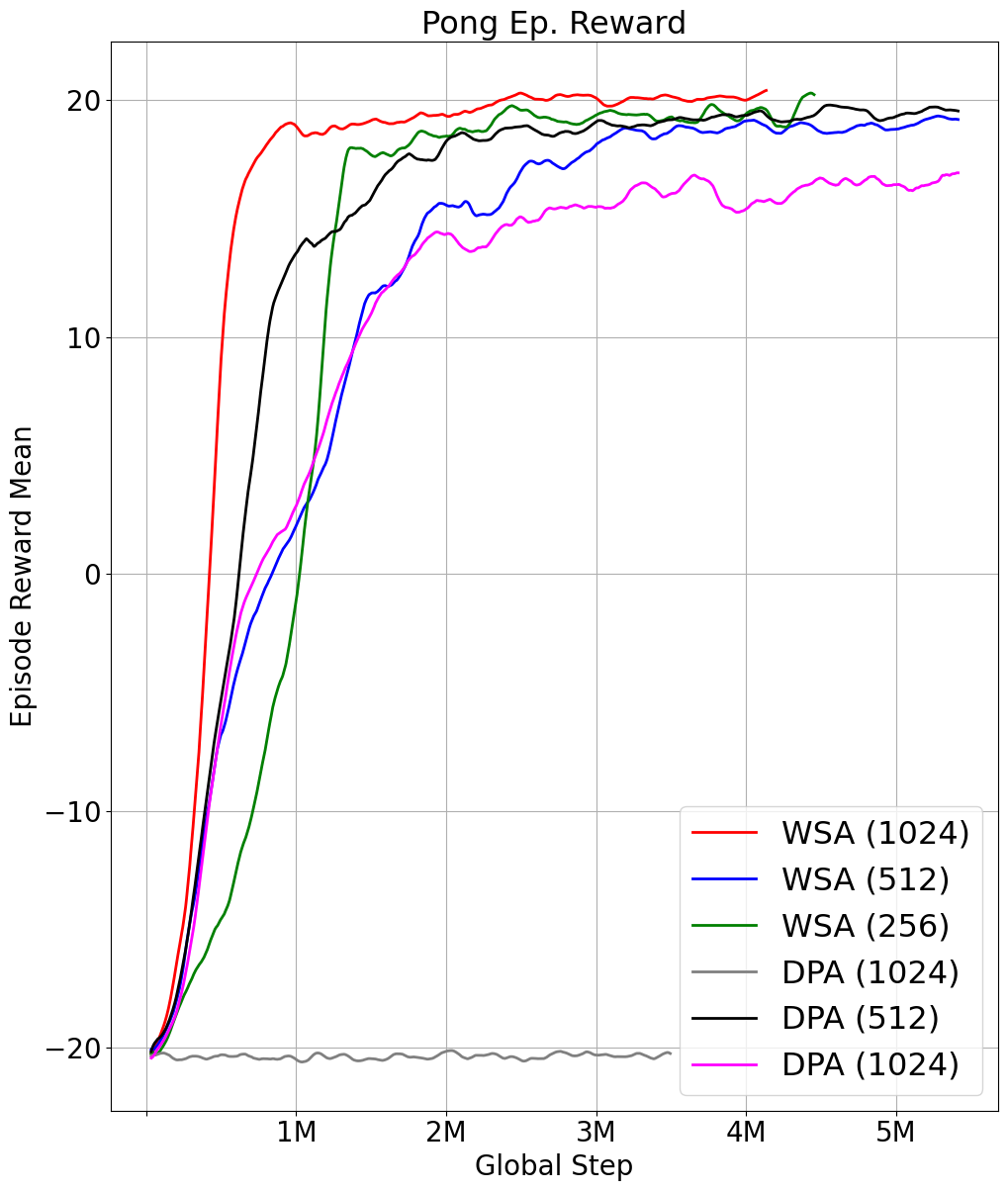}
        \caption{\texttt{Weight Sharing Attention} and \texttt{Dot Product Attention}}
        \label{fig:pong_wsa_dpa}
    \end{subfigure}
    \hfill
    \begin{subfigure}[b]{0.45\textwidth}
        \centering
        \includegraphics[width=\textwidth]{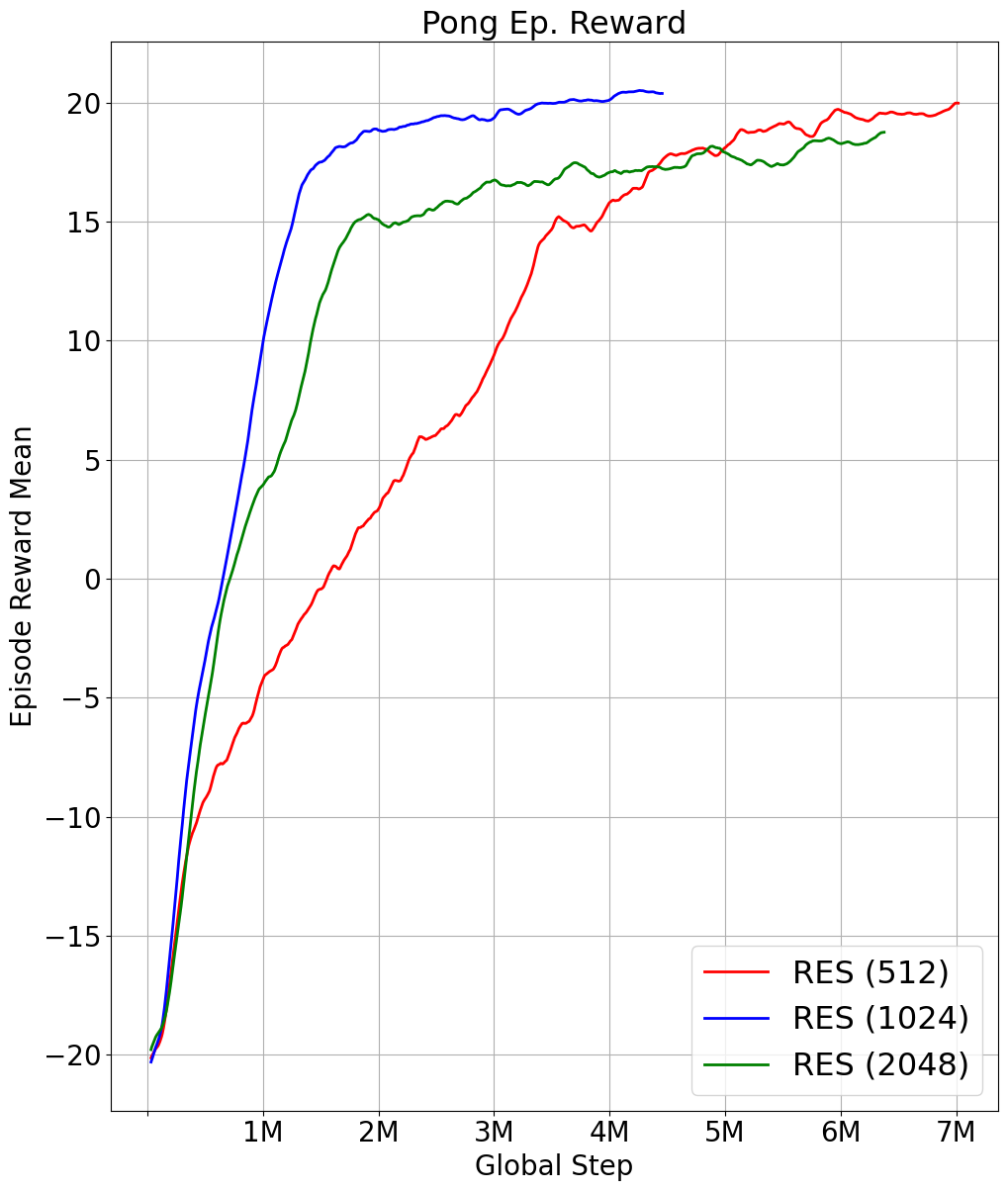}
        \caption{\texttt{Reservoir}}
        \label{fig:pong_res}
    \end{subfigure}
    \caption{Initial analysis between different combination modules configurations in \texttt{Pong}.}
    \label{fig:pong_concat_modules}
\end{figure}

\begin{figure}[ht]
    \centering
    \begin{subfigure}[b]{0.45\textwidth}
        \centering
        \includegraphics[width=\textwidth]{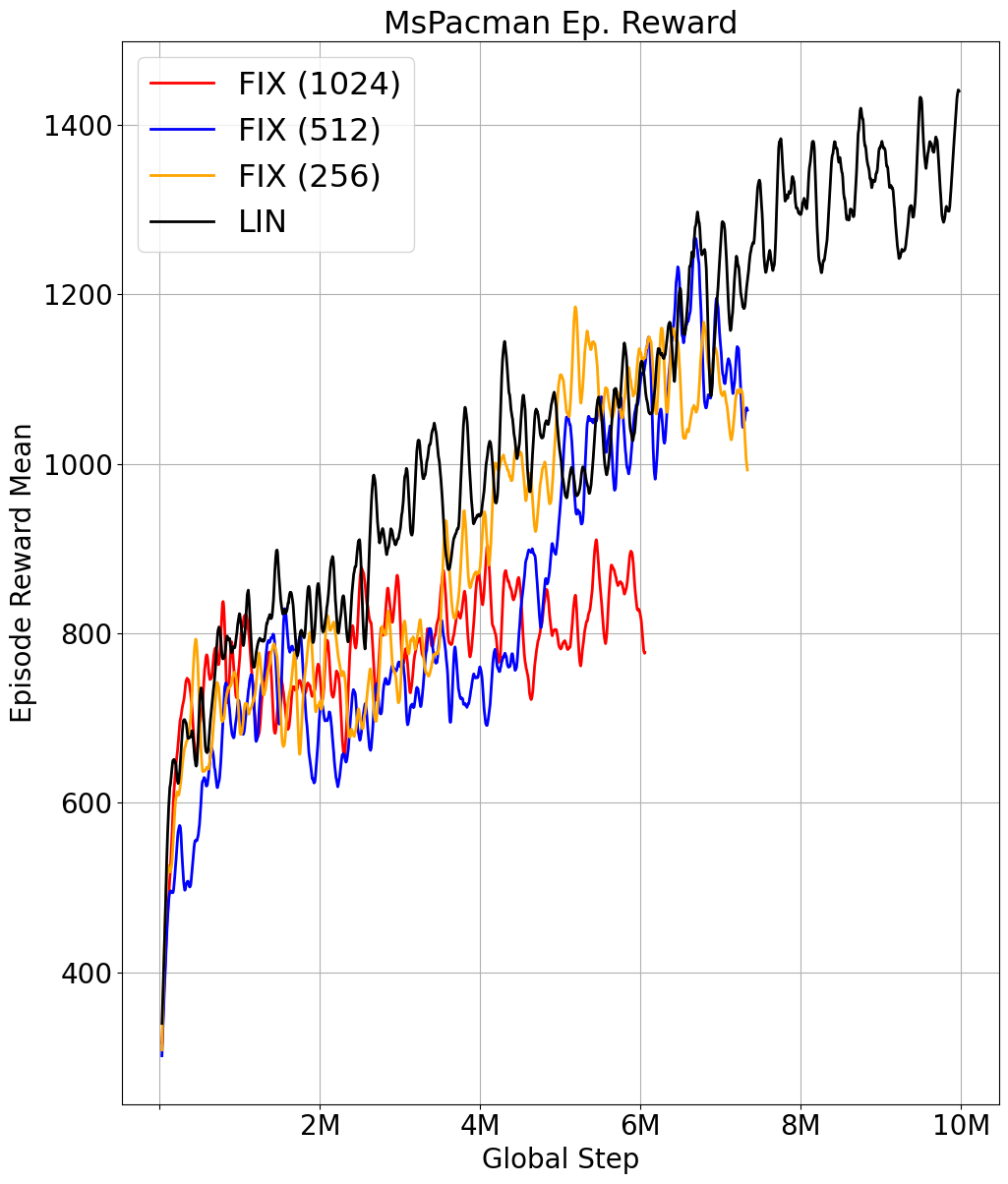}
        \caption{\texttt{Linear} and \texttt{Fixed Linear}}
        \label{fig:mspacman_lin_fix}
    \end{subfigure}
    \hfill
    \begin{subfigure}[b]{0.45\textwidth}
        \centering
        \includegraphics[width=\textwidth]{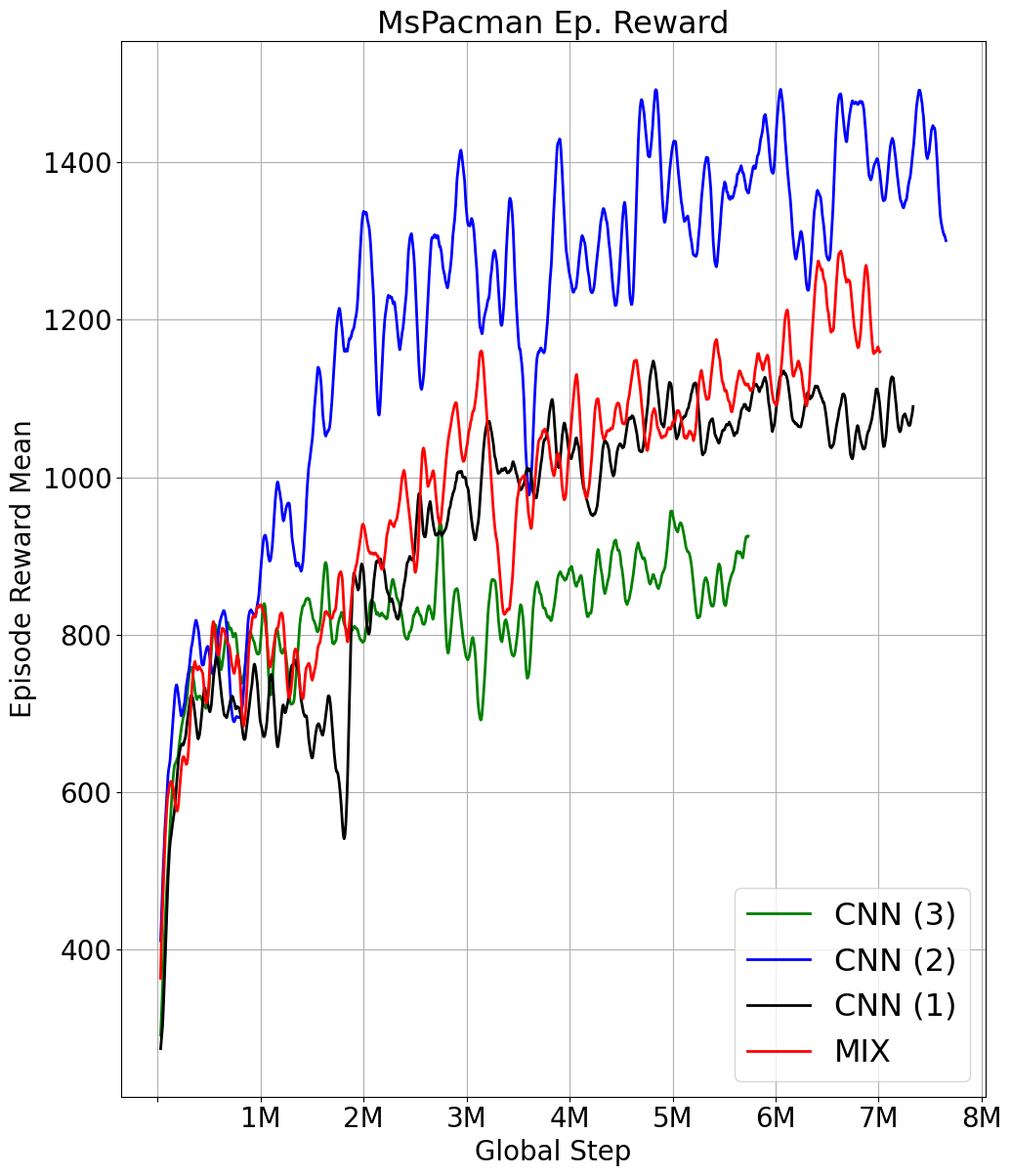}
        \caption{\texttt{Convolutional} and \texttt{Mixed}}
        \label{fig:mspacman_cnn_mixed}
    \end{subfigure}
    \hfill
    \begin{subfigure}[b]{0.45\textwidth}
        \centering
        \includegraphics[width=\textwidth]{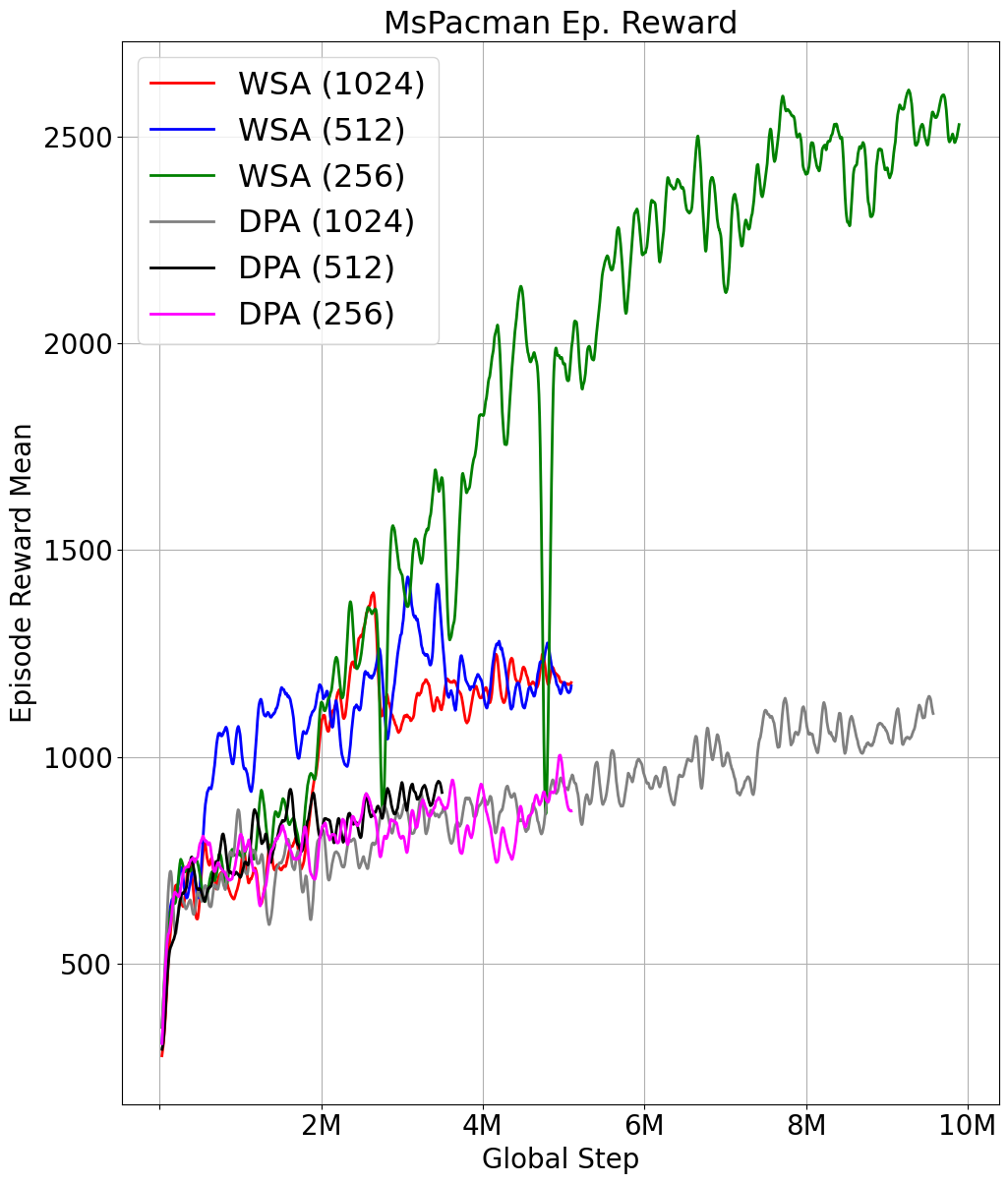}
        \caption{\texttt{Weight Sharing Attention} and \texttt{Dot Product Attention}}
        \label{fig:mspacman_wsa_dpa}
    \end{subfigure}
    \hfill
    \begin{subfigure}[b]{0.45\textwidth}
        \centering
        \includegraphics[width=\textwidth]{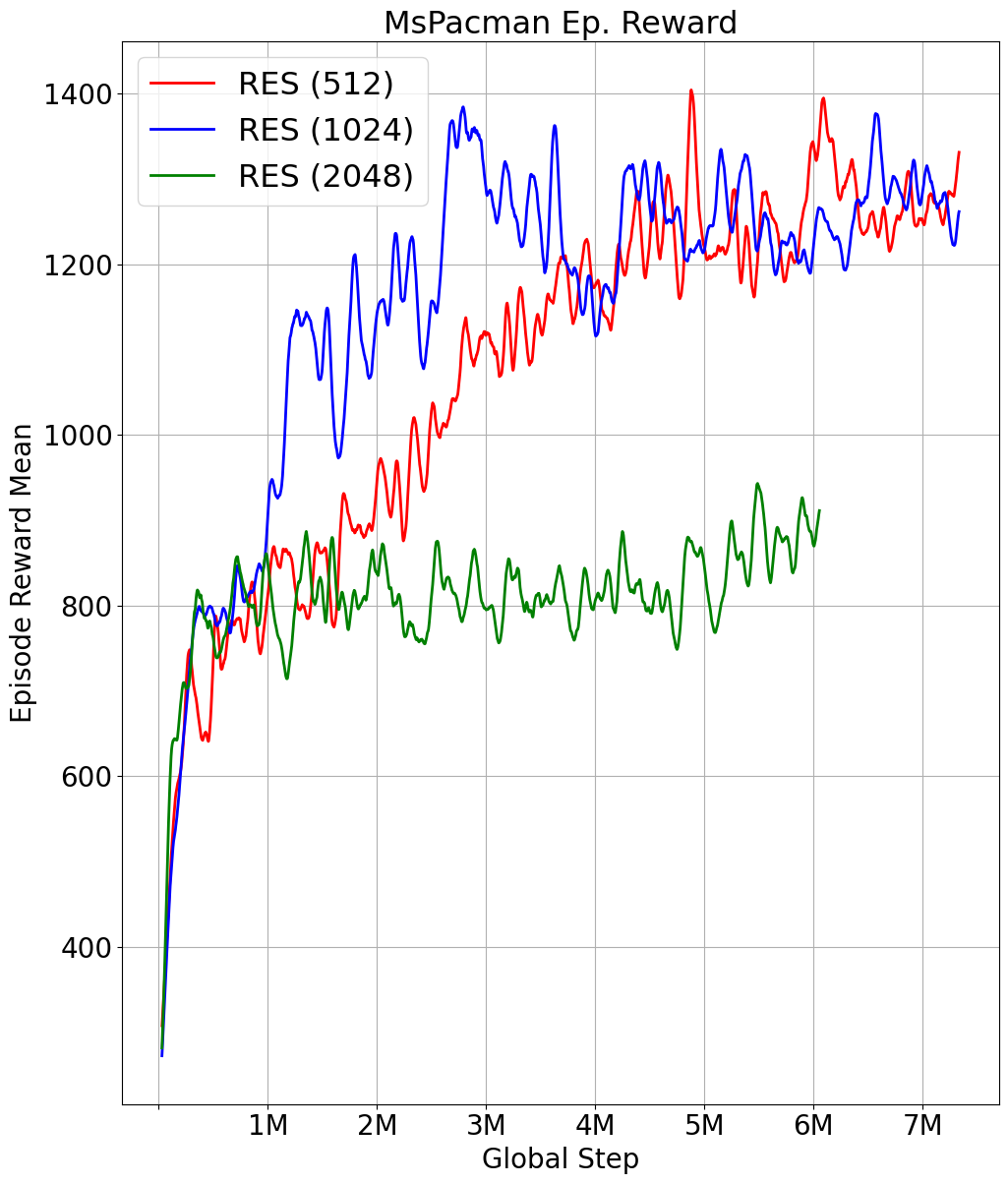}
        \caption{\texttt{Reservoir}}
        \label{fig:mspacman_res}
    \end{subfigure}
    \caption{Initial analysis between different combination modules configurations in \texttt{Ms.Pacman}.}
    \label{fig:mspacman_concat_modules}
\end{figure}

\begin{figure}[ht]
    \centering
    \begin{subfigure}[b]{0.45\textwidth}
        \centering
        \includegraphics[width=\textwidth]{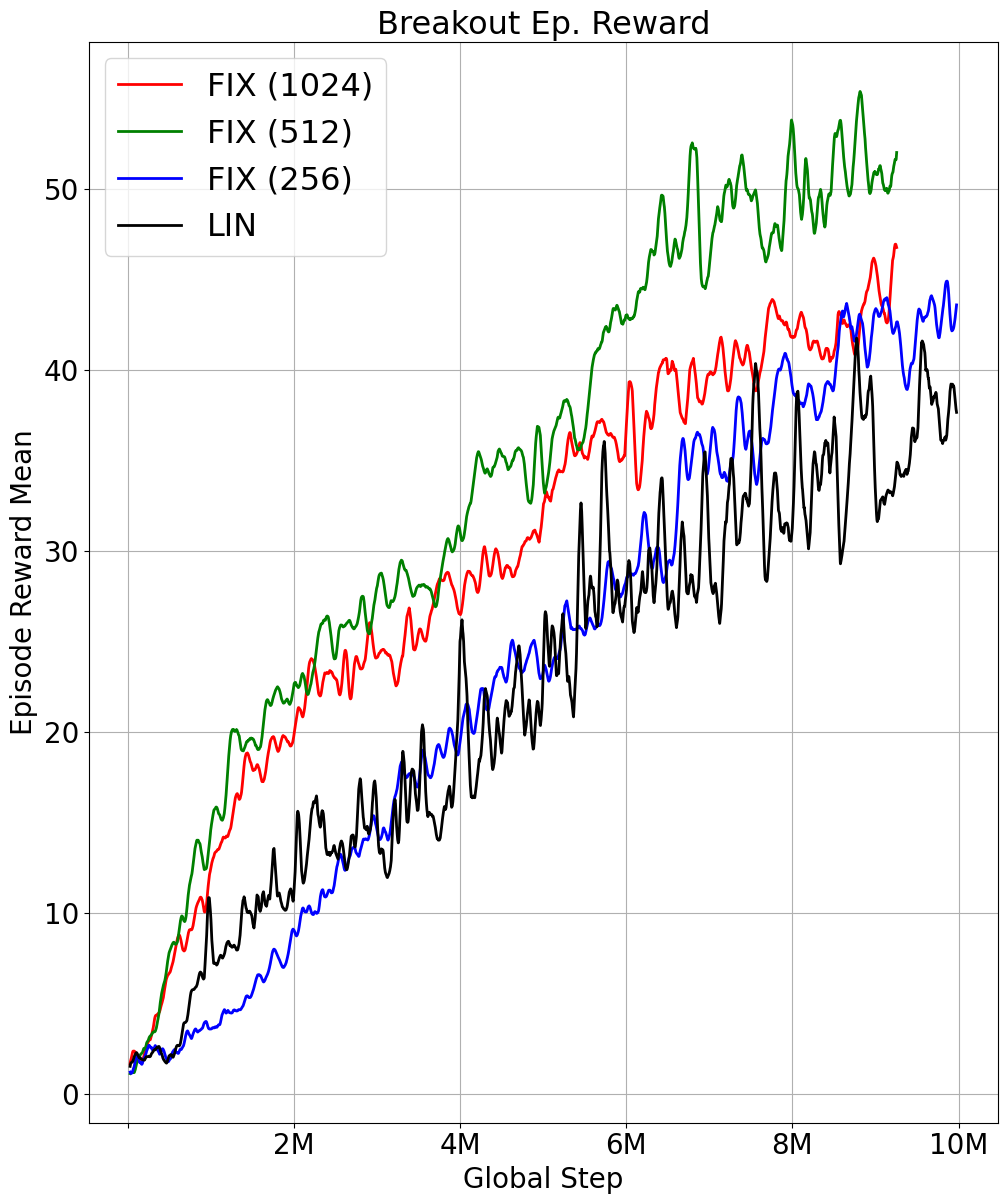}
        \caption{\texttt{Linear} and \texttt{Fixed Linear}}
        \label{fig:breakout_lin_fix}
    \end{subfigure}
    \hfill
    \begin{subfigure}[b]{0.45\textwidth}
        \centering
        \includegraphics[width=\textwidth]{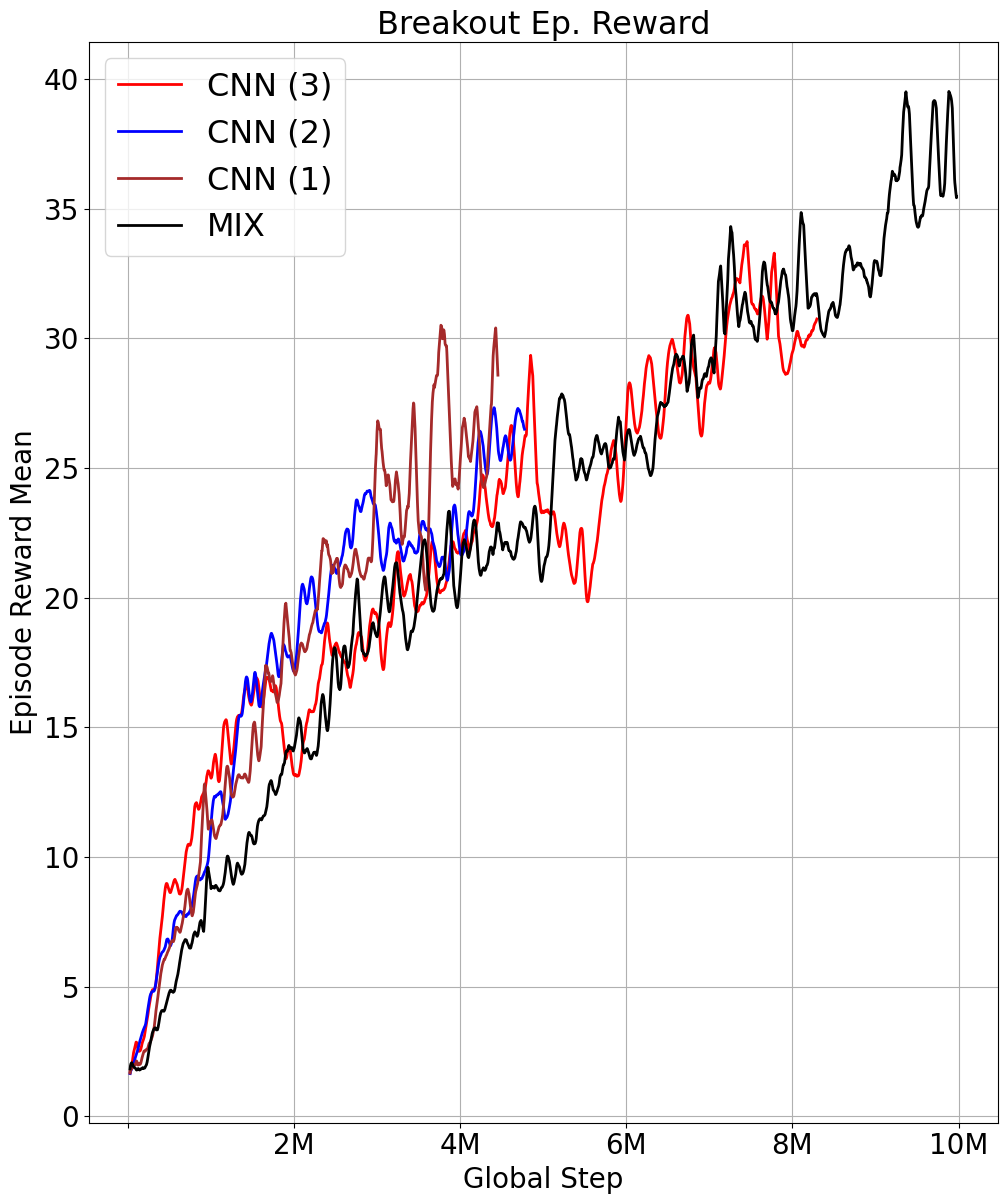}
        \caption{\texttt{Convolutional} and \texttt{Mixed}}
        \label{fig:breakout_cnn_mix}
    \end{subfigure}
    \hfill
    \begin{subfigure}[b]{0.45\textwidth}
        \centering
        \includegraphics[width=\textwidth]{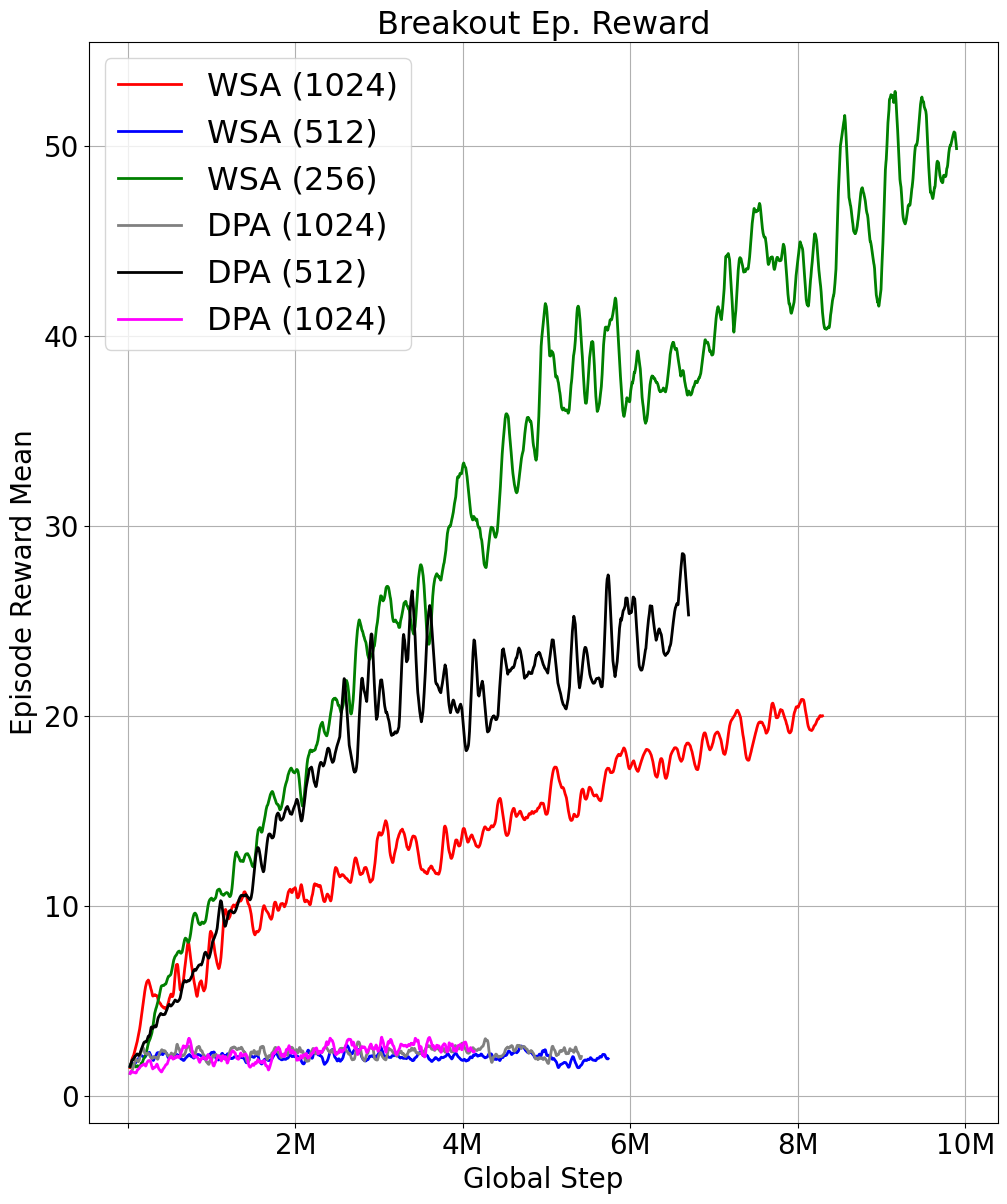}
        \caption{\texttt{Weight Sharing Attention} and \texttt{Dot Product Attention}}
        \label{fig:breakout_wsa_dpa}
    \end{subfigure}
    \hfill
    \begin{subfigure}[b]{0.45\textwidth}
        \centering
        \includegraphics[width=\textwidth]{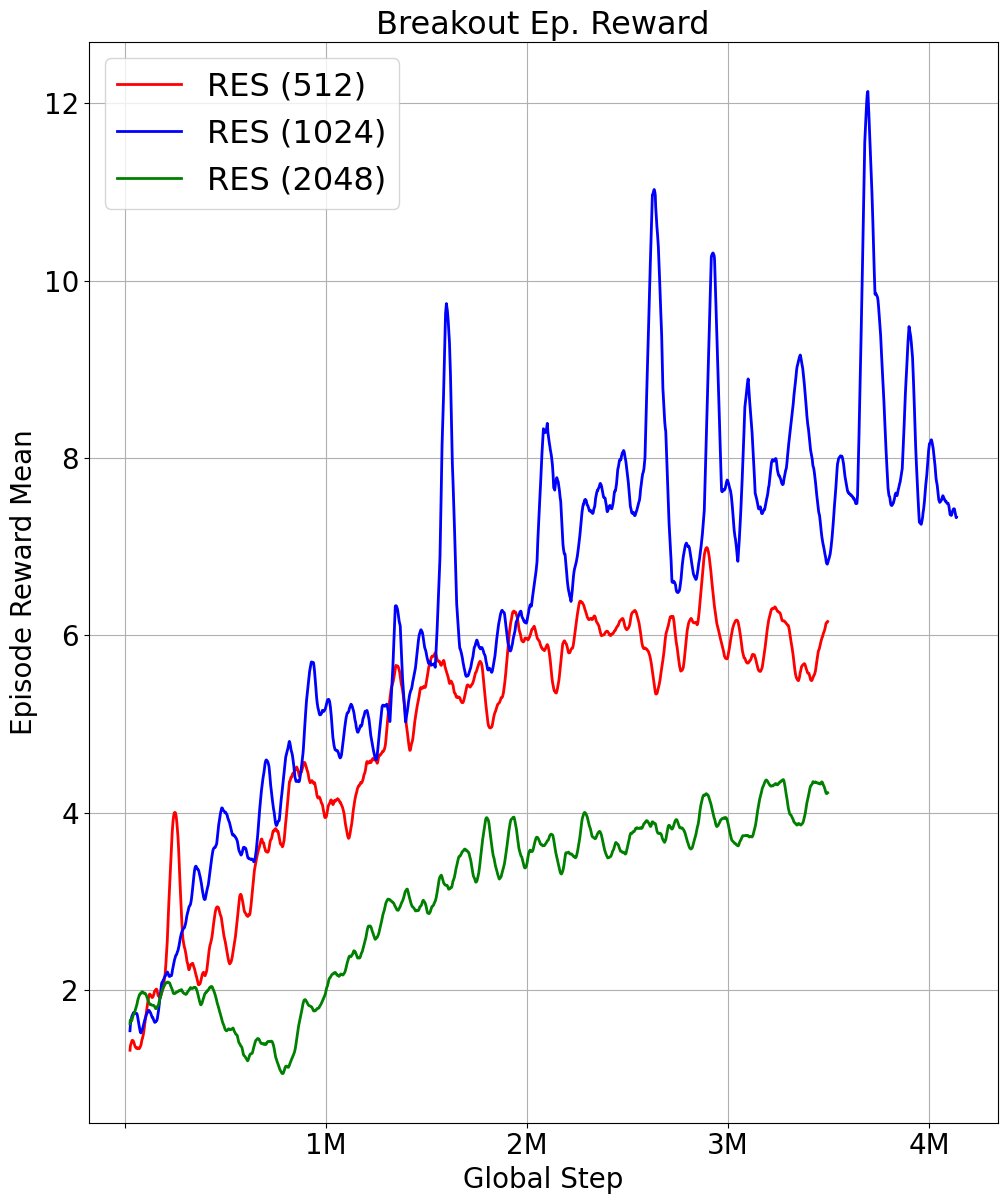}
        \caption{\texttt{Reservoir}}
        \label{fig:breakout_res}
    \end{subfigure}
    \caption{Initial analysis between different combination modules configurations in \texttt{Breakout}.}
    \label{fig:breakout_concat_modules}
\end{figure}

As shown in Table \ref{tab:emb_siz_modules}, we explore a variety of configurations for each module; visual representations are depicted in Figures \ref{fig:com1},\ref{fig:com2}, ensuring a comprehensive assessment of their performance. Table \ref{tab:parameters} reports the number of trainable parameter for each configuration. Table \ref{tab:time2} presents the time needed to compute the embeddings for each pre-trained model used by WSA in our empirical evaluation. Tables \ref{tab:time1_train}-\ref{tab:time1_eval} report respectively the timings during training and evaluation of different architectures. The E2E baseline, which uses the NatureCNN architecture \citep{mnih2013playing} (three convolutional layers and a linear layer), achieves the shortest average execution time, as expected. ENS and WSA show similar performance, with ENS being slightly faster due to the simpler combination of pre-trained model embeddings. The \textbf{computational overhead} introduced by WSA, mainly for weight computation and representation, is \textit{minimal} and \textit{comparable} to the ensemble approach, which only performs a concatenation. Moreover, the weight computation via the shared network and the construction of the final representation are efficient, requiring only a \textit{fraction of a millisecond} more than the simple concatenation operation used in ENS. Nevertheless, WSA delivers a meaningful and informative representation that leads to improved agents performance, as shown in Table \ref{tab:results} (WSA vs ENS). For both methods, the majority of the time is spent on model inference. In fact, the primary bottleneck lies in the pre-trained models themselves, as shown in Table \ref{tab:time2}. The inference of the models takes on average \textit{3.423 ms} that corresponds to \textbf{87\%} of the time needed by WSA to encode observations. The pre-trained models are an arbitrary and replaceable design choice, independent of the focus of our work. To further prove this point, C1, which uses a single SwinTransformer model \citep{liu2021swin}, is the slowest due to its high computational cost. Finally, Figures \ref{fig:pong_concat_modules}, \ref{fig:mspacman_concat_modules}, and \ref{fig:breakout_concat_modules} illustrate the training performance of each agent, offering a visual representation of their learning trajectories. Early stopping was applied for agents that showed no improvement over multiple evaluations. This strategy allowed us to promptly identify the best-performing configurations.

\begin{table}[ht] 
    \begin{center}
        \begin{tabular}{ll}
            \multicolumn{1}{l}{\bf Feature Extractor}  &\multicolumn{1}{l}{\bf Embedding Size}
            \\ \hline \\
            Linear Concatenation              &  - \\
            Fixed Linear Concatenation        & 256, 512, 1024 \\
            Convolutional Concatenation       & 1, 2, 3\\
            Mixed                             & - \\
            Reservoir Concatenation           & 512, 1024, 2048 \\
            Dot Product Attention             & 256, 512, 1024 \\
            Weights Sharing Attention         & 256, 512, 1024 \\
            
        \end{tabular}
    \end{center}
    \caption{All the extractors' configurations we tested in the initial phase of experiments. For Fixed Linear, Dot Product Attention, and Weights Sharing Attention, the values indicate the fixed dimensions of embeddings and context; for Reservoir, the size of the reservoir; and for Convolutional, the number of convolutional layers.}
    \label{tab:emb_siz_modules}
\end{table}

\begin{table}[ht]
    \begin{center}
        \begin{tabular}{ll}
            \multicolumn{1}{l}{\textbf{Feature Extractor}}  &\multicolumn{1}{l}{\textbf{Parameters}}
            \\ \hline \\
            Linear Concatenation & \texttt{8.7M} \\
            Fixed Linear Concatenation & \texttt{4.9M}--\texttt{19.4M} \\
            Convolutional Concatenation & \texttt{4.2M} \\
            Mixed & \texttt{4.5M} \\
            Reservoir Concatenation & \texttt{0.3M}--\texttt{1.1M} \\
            Dot Product Attention & \texttt{8.7M}--\texttt{34.7M} \\
            Weights Sharing Attention & \texttt{8.7M}--\texttt{34.7M} \\
            End-to-end & \texttt{1.6M} \\
        \end{tabular}
    \end{center}
    \caption{Number of lear parameters, from the smallest to the biggest configuration, based on the embedding size presented in Tab. \ref{tab:emb_siz_modules}.}
    \label{tab:parameters}
\end{table}

\begin{table}[h!]
    \centering
    \begin{tabular}{lc}
        \toprule
       \textbf{Model} &\textbf{Inference Time (ms)} \\
        \midrule
        State-Representation \citep{anand2019unsupervised}      & 0.812 $\pm$ 0.082 \\
        Object Key-Points Encoder \citep{kulkarni2019unsupervised} & 0.767 $\pm$ 0.121 \\
        Object Key-Points KeyNet \citep{jakab2018keynet, kulkarni2019unsupervised}  & 0.757 $\pm$ 0.095 \\
        Video Object Segmentation \citep{goel2018unsupervised} & 0.819 $\pm$ 0.126 \\
        Autoencoder                & 0.268 $\pm$ 0.028 \\
        \addlinespace
        \textit{Average Total}     & \textit{3.423 ms} \\
        \bottomrule
    \end{tabular}
    \caption{Inference timings of pre-trained models used in our work.}
    \label{tab:time2}
\end{table}

\begin{table}[h!]
    \centering
    \begin{tabular}{l|cccc}
        \toprule
        \textbf{Step} & \textbf{E2E} & \textbf{ENS} & \textbf{WSA} & \textbf{C1}\\
        \midrule
        \textit{Encode Observations} & 0.5129 $\pm$ 0.0561 & 3.7425 $\pm$ 0.5268 & 3.9020 $\pm$ 0.4017 & 9.6541 $\pm$ 1.0187 \\
        \textit{Policy Network} & 0.1321 $\pm$ 0.0103 & 0.1191 $\pm$ 0.0140 & 0.1113 $\pm$ 0.0073 & 0.1128 $\pm$ 0.0122 \\
        \bottomrule
    \end{tabular}
    \caption{Training timings breakdown (in \textit{ms}) of the two main step comparing different architectures.}
    \label{tab:time1_train}
\end{table}

\begin{table}[h!]
    \centering
    \begin{tabular}{l|cccc}
        \toprule
        \textbf{Step} & \textbf{E2E} & \textbf{ENS} & \textbf{WSA} & \textbf{C1}\\
        \midrule
        \textit{Encode Observations} & 0.3896 $\pm$ 0.0670 & 3.6156 $\pm$ 0.3875 & 3.9071 $\pm$ 0.4153 & 10.9849 $\pm$ 1.1776 \\
        \textit{Policy Network} & 0.1127 $\pm$ 0.0049 & 0.1162 $\pm$ 0.0084 & 0.1120 $\pm$ 0.0102 & 0.1191 $\pm$ 0.0101 \\
        \bottomrule
    \end{tabular}
    \caption{Evaluation timings breakdown (in \textit{ms}) of the two main step comparing different architectures.}
    \label{tab:time1_eval}
\end{table}

\pagebreak

\end{document}